\PassOptionsToPackage{table}{xcolor}
\documentclass[runningheads]{llncs}
\usepackage{eccv}
\usepackage{eccvabbrv}
\usepackage{graphicx}
\usepackage{booktabs}
\usepackage[accsupp]{axessibility}
\usepackage[pagebackref,breaklinks,colorlinks,citecolor=eccvblue]{hyperref}
\usepackage{orcidlink}
\usepackage[utf8]{inputenc} 
\usepackage[T1]{fontenc}  
\usepackage{url}    
\usepackage{booktabs} 
\usepackage{amsfonts}  
\usepackage{nicefrac}
\usepackage{microtype} 
\usepackage{adjustbox}
\usepackage{multirow} 
\usepackage{threeparttable}
\usepackage{amsmath,bm}
\usepackage{amssymb}
\usepackage{pifont}
\newcommand{\cmark}{\textcolor{blue}{\checkmark}}
\newcommand{\xmark}{\textcolor{black}{\ding{55}}}
\usepackage{enumitem}
\usepackage{setspace}     
\usepackage{subcaption}
\usepackage{wrapfig}
\usepackage{csquotes}
\usepackage{indentfirst}
\begin{document}

\title{BK-SDM: A Lightweight, Fast, and Cheap Version of Stable Diffusion} 

\author{Bo-Kyeong Kim$^{1}$\orcidlink{0000-0002-0224-7985} \and Hyoung-Kyu Song$^{2}$\orcidlink{0000-0002-6546-9593} \and
Thibault Castells$^{1}$\orcidlink{0009-0004-0549-5695} \and \\ 
Shinkook Choi$^{1}$\orcidlink{0000-0002-9617-2418}}

\authorrunning{B.-K.~Kim~\etal}

\institute{$^{1}$Nota Inc.\quad$^{2}$Captions Research\\
\email{\{bokyeong.kim, thibault, shinkook.choi\}@nota.ai, kyu@captions.ai}}

\maketitle

\begin{figure*}[h]
  \centering
    \includegraphics[width=\linewidth]{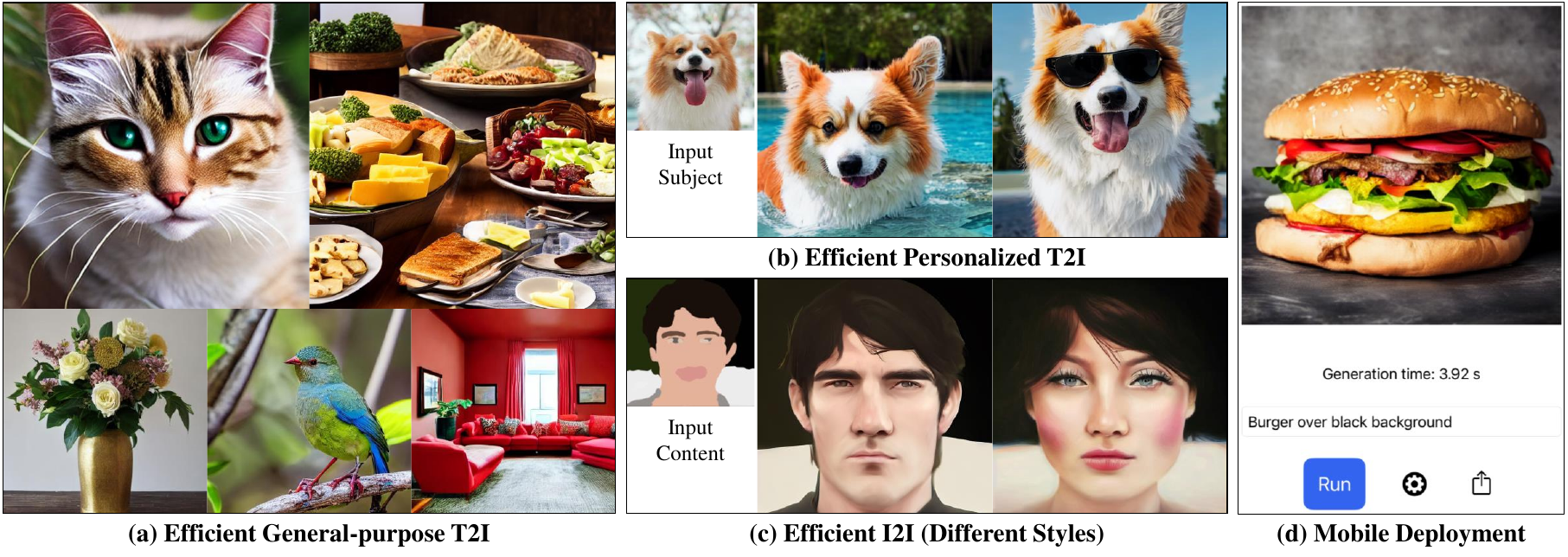}
  \caption{Our compressed model enables efficient (a) zero-shot text-to-image generation, (b) personalized synthesis, (c) image-to-image translation, and (d) mobile deployment. Samples from BK-SDM-Small with 36\% reduced parameters and latency are shown.} \label{fig:teaser}
\end{figure*}

\begin{abstract}
Text-to-image (T2I) generation with Stable Diffusion models (SDMs) involves high computing demands due to billion-scale parameters. To enhance efficiency, recent studies have reduced sampling steps and applied network quantization while retaining the original architectures. The lack of architectural reduction attempts may stem from worries over expensive retraining for such massive models. In this work, we uncover the surprising potential of block pruning and feature distillation for low-cost general-purpose T2I. By removing several residual and attention blocks from the U-Net of SDMs, we achieve 30\%$\sim$50\% reduction in model size, MACs, and latency. We show that distillation retraining is effective even under limited resources: using only 13 A100 days and a tiny dataset, our compact models can imitate the original SDMs (v1.4 and v2.1-base with over 6,000 A100 days). Benefiting from the transferred knowledge, our BK-SDMs deliver competitive results on zero-shot MS-COCO against larger multi-billion parameter models. We further demonstrate the applicability of our lightweight backbones in personalized generation and image-to-image translation. Deployment of our models on edge devices attains 4-second inference. Code and models can be found at: \url{https://github.com/Nota-NetsPresso/BK-SDM}.
\end{abstract}

\section{Introduction} \label{sec:intro}

\begin{figure}[t]
  \centering
  \includegraphics[width=0.7\linewidth]{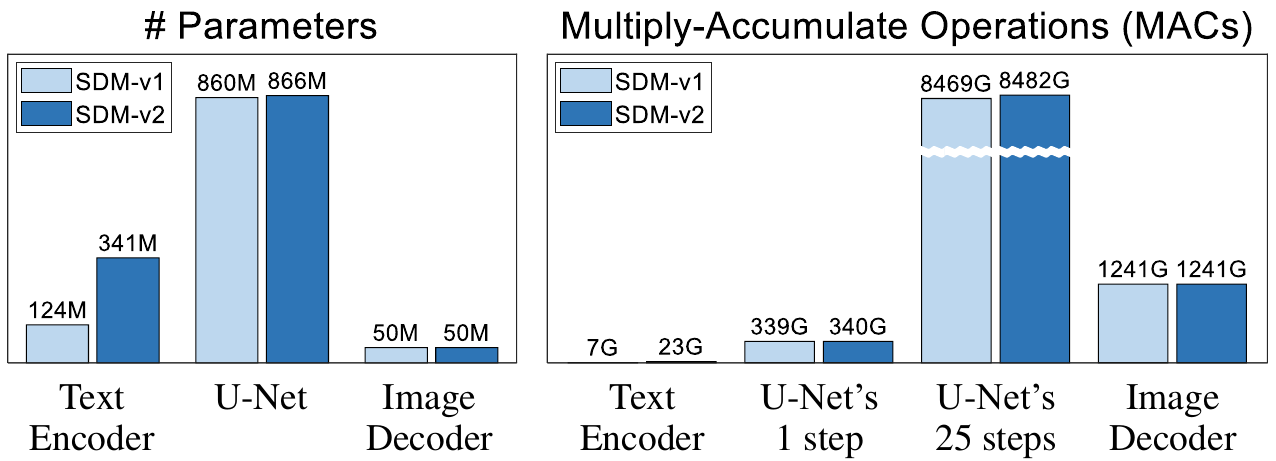}
    \caption{\textbf{Computation of Stable Diffusion}. The denoising U-Net is the main processing bottleneck. THOP~\cite{thop} is used to measure MACs in generating a 512×512 image.}\label{fig:sdm_compute}
\end{figure}

Stable Diffusion models (SDMs)~\cite{sdm_v1.4_hf,ldm2022,sdm_v1.5_hf,sdm_v2-1-base_hf} are one of the most renowned open-source models for text-to-image (T2I) synthesis, and their exceptional capability has begun to be leveraged as a backbone in several text-guided vision applications~\cite{controlnet, instructpix2pix, alignVideoGen, wang2022score}. SDMs are T2I-specialized latent diffusion models (LDMs)~\cite{ldm2022}, which employ diffusion operations~\cite{clsfreeguide,ddim,pndm} in a semantically compressed space for compute efficiency. Within a SDM, a U-Net~\cite{unet,adm} performs iterative sampling to progressively denoise a random latent code and is aided by a text encoder~\cite{radford2021learning} and an image decoder~\cite{vqgan,vqvae} to produce text-aligned images. This inference process still involves excessive computational requirements (see Fig.~\ref{fig:sdm_compute}), which often hinder the utilization of SDMs despite their rapidly growing usage. 

To alleviate this issue, numerous approaches toward efficient SDMs have been introduced. A pretrained diffusion model is distilled to reduce the number of denoising steps, enabling an identically architectured model with fewer sampling steps~\cite{onDistil_cvpr23, lcm}. Post-training quantization~\cite{li2023q,qualcomm_q,intel_q} and implementation optimization~\cite{chen2023speed,choi2023squeezing} methods are also leveraged. However, the removal of architectural elements in large diffusion models remains less explored. 

This study unlocks the immense potential of classical architectural compression in attaining smaller and faster diffusion models. We eliminate multiple residual and attention blocks from the U-Net of a SDM and retrain it with feature-level knowledge distillation (KD)~\cite{fitnets,heo2019comprehensive} for general-purpose T2I. Under restricted training resources, our compact models can mimic the original SDM by leveraging transferred knowledge. Our work effectively reduces the computation of SDM-v1.4~\cite{sdm_v1.4_hf} and SDM-v2.1-base~\cite{sdm_v2-1-base_hf} while achieving compelling zero-shot results on par with multi-billion parameter models~\cite{dalle1,cogview1,cogview2}. Our contributions are summarized as follows:
 
\begin{enumerate}

\item[$\circ$]  We compress SDMs by removing architectural blocks from the U-Net, achieving up to 51\% reduction in model size and 43\% improvement in latency on CPU and GPU. Previous pruning studies~\cite{fang2023depgraph, Yu_2023_CVPR, murti2023tvsprune} focused on small models ($<$100M parameters) like ResNet50 and DeiT-B, not on foundation models like SDMs ($>$1,000M$=$1B), possibly due to the lack of economic retraining for such large models. Moreover, U-Net architectures are arguably more complex due to the necessity of considering skip connections across the network, making the structural block removal inside them not straightforward. 

\item[$\circ$]  To the best of our knowledge, we first demonstrate the notable benefit of feature distillation for training diffusion models, which enables competitive T2I even with significantly fewer resources (using only 13 A100 days and 0.22M LAION pairs~\cite{laion_blog}). Considering the vast expense of training SDMs from scratch (surpassing 6,000 A100 days and 2,000M pairs), our study indicates that network compression is a remarkably cost-effective strategy in building compact general-purpose diffusion models.

\item[$\circ$] We show the practicality of our work across various aspects. Our lightweight backbones are readily applicable to customized generation~\cite{dreambooth} and image-to-image translation~\cite{sdedit}, effectively lowering finetuning and inference costs. T2I synthesis on Jetson AGX Orin and iPhone 14 using our models takes less than 4 seconds.

\item[$\circ$] We have publicly released our approach, model weights, and source code, motivating subsequent works by other researchers (e.g., block pruning and KD for the SDM-v1 variant~\cite{segmind_sd-v1,edgefusion} and SDXL~\cite{segmind_sdxl,koala}).

\end{enumerate}

\section{Related Work} \label{sec:relatedwork}

\noindent \textbf{Large T2I diffusion models.} 
By gradually removing noise from corrupted data, diffusion-based generative models~\cite{ddpm,ddim,adm} enable high-fidelity synthesis with broad mode coverage. Integrating these merits with the advancement of pretrained language models~\cite{radford2021learning,radford2019language,bert} has significantly improved the quality of T2I synthesis. In GLIDE~\cite{glide} and Imagen~\cite{imagen}, a text-conditional diffusion model generates a small image, which is upsampled via super-resolution modules. In DALL·E-2~\cite{dalle2}, a text-conditional prior network produces an image embedding, which is transformed into an image via a diffusion decoder and further upscaled into higher resolutions. SDMs~\cite{sdm_v1.4_hf,sdm_v1.5_hf,sdm_v2-1-base_hf,ldm2022} perform the diffusion modeling in a low-dimensional latent space constructed through a pixel-space autoencoder. We use SDMs as our baseline because of its open-access and gaining popularity over numerous downstream tasks~\cite{instructpix2pix, wang2022score, alignVideoGen, dreambooth}.

\noindent \textbf{Efficient diffusion models.} Several studies have addressed the slow sampling process. Diffusion-tailored distillation~\cite{onDistil_cvpr23, onDistil_neuripsw22,salimans2022progressive} progressively transfers knowledge from a pretrained diffusion model to a fewer-step model with the same architecture. Fast high-order solvers~\cite{dpm_solver_1,dpm_solver_2,zhang2022fast} for diffusion ordinary differential equations boost the sampling speed. Complementarily, our network compression approach reduces per-step computation and can be easily integrated with less sampling steps. Leveraging quantization~\cite{li2023q,qualcomm_q,intel_q} and implementation optimizations~\cite{chen2023speed,choi2023squeezing} for SDMs can also be combined with our compact models for further efficiency.

\noindent \textbf{Distillation-based compression.}
KD enhances the performance of small-size models by exploiting output-level~\cite{hintonKd,rkd} and feature-level~\cite{fitnets,heo2019comprehensive,at_kd} information of large source models. Although this classical KD has been actively used for efficient GANs~\cite{li2020gan,ren2021omgd,zhang2022wavelet}, its power has not been explored for structurally compressed diffusion models. Distillation pretraining enables small yet capable general-purpose language models~\cite{distilbert,mobilebert,tinybert} and vision transformers~\cite{deit,hao2022learning}. Beyond such models, we show that its success can be extended to diffusion models with iterative sampling.

\noindent \textbf{Concurrent studies.}
SnapFusion~\cite{li2023snapfusion} and MobileDiffusion~\cite{mobile-diffusion} achieve an efficient U-Net for SDMs through architecture optimization and step reduction. Würstchen~\cite{wuerstchen} introduces two diffusion processes on low- and high-resolution latent spaces for economic training. These works are valuable but require much larger training resources than our work (see Tab.~\ref{table:main_comp}). In contrast, we utilize considerably fewer resources, leveraging the surprising benefits of classical KD for foundational diffusion models. While not demonstrated on SDMs, Diff-Pruning~\cite{fang2023structural} proposes structured pruning based on Taylor expansion tailored for diffusion models. We emphasize the use of depth (block) pruning in our work, which often leads to greater enhancements in inference speeds compared to the width (channel) pruning approach of Diff-Pruning. Moreover, Tab.~\ref{table:prune_crit} employs a comparable Taylor pruning criterion, which results in an insufficient decrease in MACs.
\section{Compression Method} \label{sec:method}

We compress the U-Net~\cite{unet} in SDMs, which is the most compute-heavy component (see Fig.~\ref{fig:sdm_compute}). Conditioned on the text and time-step embeddings, the U-Net performs multiple denoising steps on latent representations. At each step, the U-Net produces the noise residual to compute the latent for the next step. We reduce this per-step computation, leading to \textit{Block-removed Knowledge-distilled SDMs} (BK-SDMs).

\subsection{Compact U-Net Architecture}

\begin{figure*}[t]
  \centering
    \includegraphics[width=\linewidth]{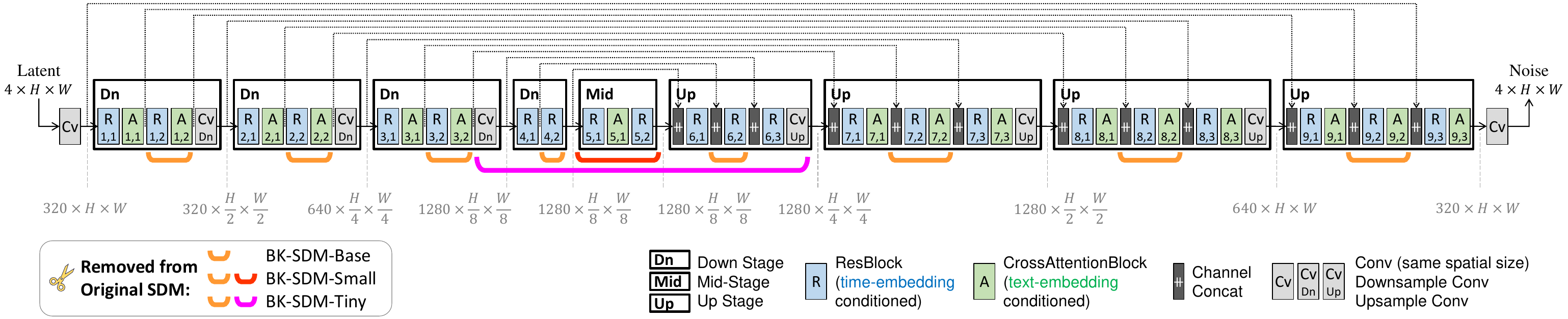}
  \caption{\textbf{Block removal from the denoising U-Net.} Our approach is applicable to all the SDM versions in v1 and v2, which share the same U-Net block configuration. For experiments, we used v1.4~\cite{sdm_v1.4_hf} and v2.1-base~\cite{sdm_v2-1-base_hf}. See Sec.~\ref{appendix:model_details} for the details.} \label{fig_arch}
\end{figure*}

The following architectures are obtained by compressing SDM-v1 (1.04B parameters), as shown in Fig.~\ref{fig_arch}:

\begin{enumerate}
\item[$\circ$] BK-SDM-Base (0.76B) obtained with Sec.~\ref{method_fewerBlk}.(1).
\item[$\circ$] BK-SDM-Small (0.66B) with Secs.~\ref{method_fewerBlk}.(1)~and~\ref{method_midrm}.(2).
\item[$\circ$] BK-SDM-Tiny (0.50B) with Secs.~\ref{method_fewerBlk}.(1),~\ref{method_midrm}.(2),~and~\ref{method_innerrm}.(3).
\end{enumerate}

\noindent Our approach can be identically applied to SDM-v2 (1.26B parameters), leading to BK-SDM-v2-\{Base (0.98B), Small (0.88B), Tiny (0.72B)\}.

\subsubsection{(1) Fewer Blocks in the Down and Up Stages.} \label{method_fewerBlk}

This approach is closely aligned with DistilBERT~\cite{distilbert} which halves the number of layers and initializes the compact model with the original weights by benefiting from the shared dimensionality. In the original U-Net, each stage with a common spatial size consists of multiple blocks, and most stages contain pairs of residual (R)~\cite{he2016deep} and cross-attention (A)~\cite{vaswani2017attention,jaegle2021perceiver} blocks. We hypothesize the existence of some unnecessary pairs and use the following removal strategies. 

\begin{enumerate}

\item[$\circ$] \textit{Down Stages.} We maintain the first R-A pairs while eliminating the second pairs, because the first pairs process the changed spatial information and would be more important than the second pairs. This design is consistent with the sensitivity analysis that measures the block-level significance (see Fig.~\ref{fig:prune_sensitivity_anal}). Our approach also does not harm the dimensionality of the original U-Net, enabling the use of the corresponding pretrained weights for initialization~\cite{distilbert}.

\item[$\circ$] \textit{Up Stages.} While adhering to the aforementioned scheme, we retain the third R-A pairs. This allows us to utilize the output feature maps at the end of each down stage and the corresponding skip connections between the down and up stages. The same process is applied to the innermost down and up stages that contain only R blocks. 

\end{enumerate}

\begin{table*}[t]
\centering\small
\begin{minipage}{.29\textwidth}
\centering\small
\includegraphics[width=\linewidth]{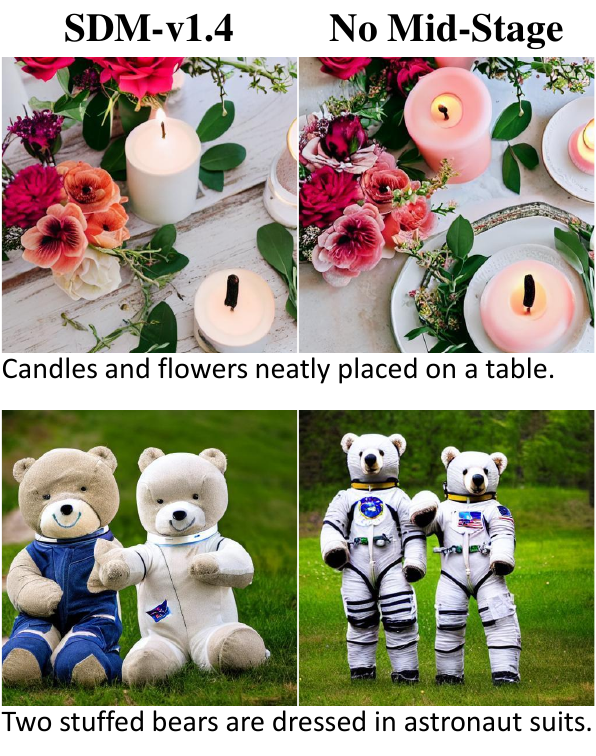}
\captionof{figure}{\textbf{Minor impact of removing the mid-stage from the U-Net.} Results without retraining. See Sec.~\ref{appendix:mid_removal} for additional results.} \label{fig_mid_rm}
\end{minipage}
~~~
\hspace{-0.3cm}
\begin{minipage}{.68\textwidth}
\centering\small
\includegraphics[width=\textwidth]{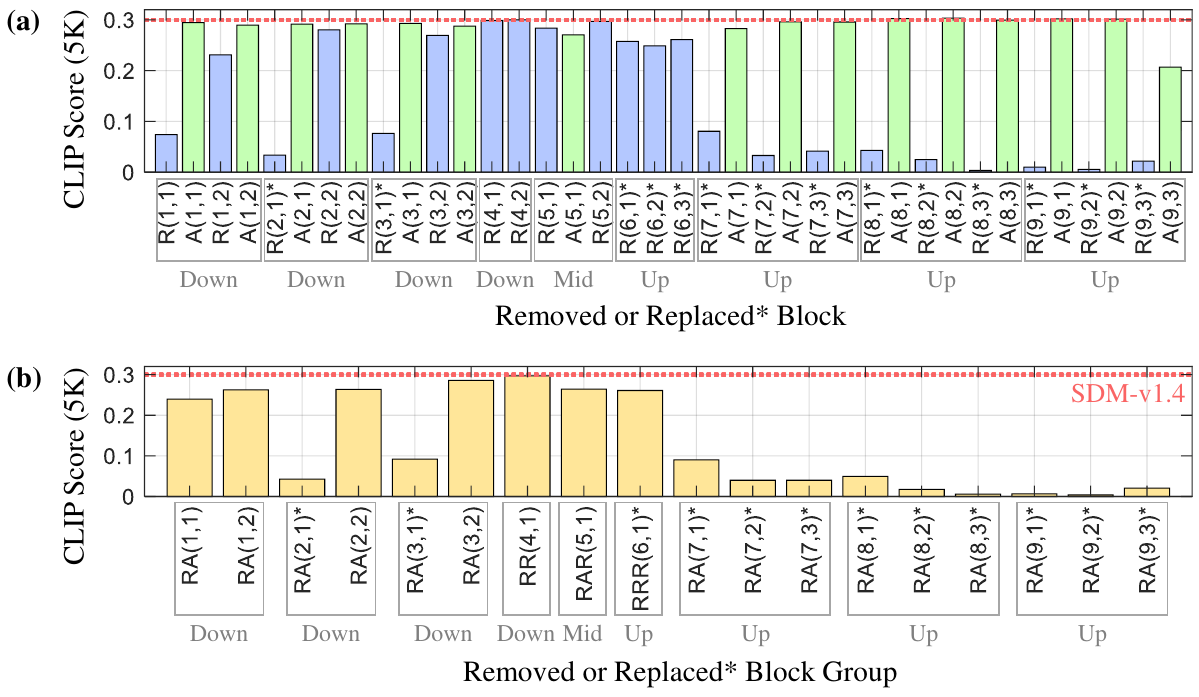}
\captionof{figure}{\textbf{Importance of (a) each block and (b) each group of paired/triplet blocks.} Higher score implies removable blocks. The results are aligned with our architectures (e.g., removal of innermost stages and the second R-A pairs in down stages). See Sec.~\ref{appendix:prune_sens_anal} for further analysis.}\label{fig:prune_sensitivity_anal}
\end{minipage}
\end{table*}

\subsubsection{(2) Removal of the Entire Mid-Stage.} \label{method_midrm}

Surprisingly, removing the entire mid-stage from the original U-Net does not noticeably degrade the generation quality while effectively reducing the parameters by 11\% (see Fig.~\ref{fig_mid_rm}). This observation is consistent with the minor role of inner layers in the U-Net generator of GANs~\cite{kim2022cut}.

Integrating the mid-stage removal with fewer blocks in Sec.~\ref{method_fewerBlk} further decreases compute burdens (Tab.~\ref{table:compute}) at the cost of a slight decline in performance (Tab.~\ref{table:main_comp}). Therefore, we offer this mid-stage elimination as an option, depending on the priority between compute efficiency (using BK-SDM-Small) and generation quality (BK-SDM-Base).

\subsubsection{(3) Further Removal of the Innermost Stages.} \label{method_innerrm}

For additional compression, the innermost down and up stages can also be pruned, leading to our lightest model BK-SDM-Tiny. This implies that outer stages with larger spatial dimensions and their skip connections play a crucial role in the U-Net for T2I synthesis.

\subsubsection{Alignment with Pruning Sensitivity Analysis.}
To support the properness of our architectures, we measure the importance of each block (see Fig.~\ref{fig:prune_sensitivity_anal}) and show that unimportant blocks match with our design choices. The importance is measured by how generation scores vary when removing each residual or attention block from the U-Net. A significant drop in performance highlights the essential role of that block. Note that some blocks are not directly removable due to different channel dimensions between input and output; we replace such blocks with channel interpolation modules (denoted by “*” in Fig.~\ref{fig:prune_sensitivity_anal}) to mimic the removal while retaining the information.

The sensitivity analysis implies that the innermost down-mid-up stages and the second R-A pairs in the down stages play relatively minor roles. Pruning these blocks aligns with our architectures, designed based on human knowledge (e.g., prioritizing blocks with altered channel dimensions) and previous studies~\cite{distilbert, kim2022cut}. 

Though some results aligned, the importance criterion from the CLIP Score pruning sensitivity (in Fig.~\ref{fig:prune_sensitivity_anal}) leads to excessive pruning of attention blocks, eventually yielding inferior performance compared to our approach (see Table~\ref{table:prune_crit}).

\begin{figure}[t]
  \centering
  \includegraphics[width=0.6\linewidth]{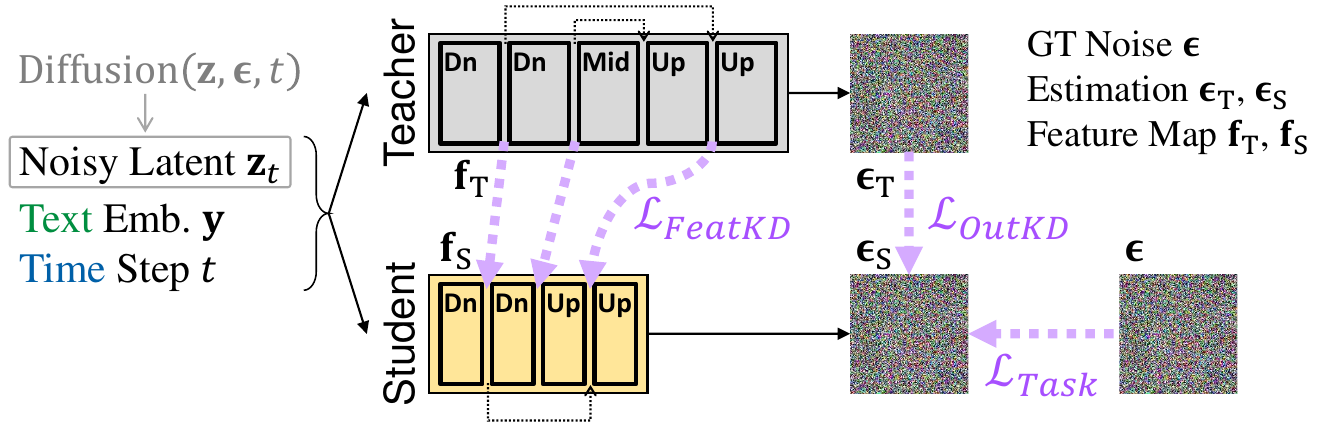}
    \caption{\textbf{Distillation-based retraining.} The block-removed U-Net is trained effectively through the guidance of the original U-Net.}\label{fig:kd}
\end{figure}

\subsection{Distillation-based Retraining}

For general-purpose T2I, we train our block-removed U-Net to mimic the behavior of the original U-Net (see Fig.~\ref{fig:kd}). To obtain the input of U-Net, we use pretrained-and-frozen encoders~\cite{ldm2022} for images and text prompts.

Given the latent representation $\mathbf{z}$ of an image and its paired text embedding $\mathbf{y}$, the task loss for the reverse denoising process~\cite{ddpm,ldm2022} is computed as:

\begin{equation} \label{loss_task}
\mathcal{L}_{\mathrm{Task}} = \mathbb{E}_{\mathbf{z}, \bm{\mathrm{\epsilon}}, \mathbf{y}, t} \Big[ ||\bm{\mathrm{\epsilon}}-\epsilon_{\mathrm{S}}(\mathbf{z}_t, \mathbf{y}, t)||_2^2 \Big],
\end{equation}

\noindent where $\mathbf{z}_t$ is a noisy latent code from the diffusion process~\cite{ddpm} with the sampled noise $\bm{\mathrm{\epsilon}}$$\sim$$N(\mathbf{0},\mathbf{I})$ and time step $t$$\sim$$\mathrm{Uniform}(1,T)$, and $\epsilon_{\mathrm{S}}(\circ)$ indicates the estimated noise from our compact U-Net student. For brevity, we omit the subscripts of $\mathbb{E}_{\mathbf{z}, \bm{\mathrm{\epsilon}}, \mathbf{y}, t} [ \circ ]$ in the following notations. 

The compact student is also trained to imitate the outputs of the original U-Net teacher, $\epsilon_{\mathrm{T}}(\circ)$, with the following output-level KD objective~\cite{hintonKd}:

\begin{equation} \label{loss_out_kd}
\mathcal{L}_{\mathrm{OutKD}} = \mathbb{E}\Big[ ||\epsilon_{\mathrm{T}}(\mathbf{z}_t, \mathbf{y}, t)-\epsilon_{\mathrm{S}}(\mathbf{z}_t, \mathbf{y}, t)||_2^2 \Big].
\end{equation}

A key to our approach is feature-level KD~\cite{fitnets,heo2019comprehensive} that provides abundant guidance for the student's training:

\begin{equation} \label{loss_feat_kd}
\mathcal{L}_{\mathrm{FeatKD}} = \mathbb{E}\Big[ \sum_{l}||f_{\mathrm{T}}^l(\mathbf{z}_t, \mathbf{y}, t)-f_{\mathrm{S}}^l(\mathbf{z}_t, \mathbf{y}, t)||_2^2 \Big],
\end{equation}

\noindent where $f_{\mathrm{T}}^l(\circ)$ and $f_{\mathrm{S}}^l(\circ)$ represent the feature maps of the $l$-th layer in a predefined set of distilled layers from the teacher and the student, respectively. While learnable regressors (e.g., 1×1 convolutions to match the number of channels) have been commonly used~\cite{shu2021channel,ren2021omgd,fitnets}, our approach circumvents this requirement. By applying distillation at the end of each stage in both models, we ensure that the dimensionality of the feature maps already matches, thus eliminating the need for additional regressors.

The final objective is shown below, and we simply set $\lambda_{\mathrm{OutKD}}$ and $\lambda_{\mathrm{FeatKD}}$ as 1. Without loss-weight tuning, our approach is effective in empirical validation.

\begin{equation} \label{loss_final}
\mathcal{L} = \mathcal{L}_{\mathrm{Task}} + \lambda_{\mathrm{OutKD}}\mathcal{L}_{\mathrm{OutKD}} + \lambda_{\mathrm{FeatKD}} \mathcal{L}_{\mathrm{FeatKD}}.
\end{equation}

\section{Experimental Setup}\label{sec:setup}

\noindent \textbf{Distillation Retraining.} We primarily use 0.22M image-text pairs from LAION-Aesthetics V2 (L-Aes) 6.5+~\cite{laion_blog, laion5b}, which are significantly fewer than the original training data used for SDMs~\cite{sdm_v1.4_hf, sdm_v2-1-base_hf} (>2,000M pairs). In Fig.~\ref{fig:datasize_iter}, dataset sizes smaller than 0.22M are randomly sampled from L-Aes 6.5+, while those larger than 0.22M are from L-Aes 6.25+.

\noindent \textbf{Zero-shot T2I Evaluation.} Following the popular protocol~\cite{dalle1,ldm2022, imagen}, we use 30K prompts from the MS-COCO validation split~\cite{mscoco}, downsample the 512×512 generated images to 256×256, and compare them with the entire validation set. We compute Fréchet Inception Distance (FID)~\cite{heusel2017gans} and Inception Score (IS)~\cite{salimans2016improved} to assess visual quality. We measure CLIP score~\cite{radford2021learning, clipscore} with CLIP-ViT-g/14 model to assess text-image correspondence. 

\noindent \textbf{Downstream Tasks.} For personalized generation, we use the DreamBooth dataset~\cite{dreambooth} (30 subjects × 25 prompts × 4$\sim$6 images) and perform per-subject finetuning. Following the evaluation protocol~\cite{dreambooth}, we use ViT-S/16 model~\cite{dino} for DINO score and CLIP-ViT-g/14 model for CLIP-I and CLIP-T scores. For image-to-image translation, input images are sourced from Meng~\etal~\cite{onDistil_cvpr23}.

\noindent \textbf{Implementation Details.} We adjust the codes in Diffusers~\cite{diffusers} and PEFT~\cite{peft}. We use a single NVIDIA A100 80G GPU for main retraining and a single NVIDIA GeForce RTX 3090 GPU for per-subject finetuning. For compute efficiency, we always opt for 25 denoising steps of the U-Net at the inference phase, unless specified. The classifier-free guidance scale~\cite{clsfreeguide,imagen} is set to the default value of 7.5. The latent resolution is set to the default ($H=W=64$ in Fig.~\ref{fig_arch}), yielding 512×512 images. See Sec.~\ref{appendix:impl_details} for the details.

\section{Results} \label{sec:results}

All the results in Secs.~\ref{main_results}–\ref{sec_train_resources} were obtained with the full benchmark protocal (MS-COCO 256×256 30K samples), except for Fig.~\ref{fig:tradeoff} (512×512 5K samples). Unless specified, the training setup in Tab.~\ref{table:main_comp} was used.

\begin{table*}[t]
\centering
\caption{\textbf{Results on zero-shot MS-COCO 256×256 30K.} Training resources include image-text pairs, batch size, iterations, and A100 days. Despite far smaller resources, our compact models outperform prior studies~\cite{dalle1, cogview1, cogview2, lafite2022, galip}, showing the benefit of compressing existing powerful models. Note that FID fluctuates more than the other metrics over training progress in our experiments (see Figs.~\ref{fig:iter_scores}~and~\ref{fig:datasize_iter}).
}\label{table:main_comp}

\begin{adjustbox}{max width=0.98\linewidth}
\begin{threeparttable}
\begin{tabular}{lcc|ccc|ccc}
\specialrule{.2em}{.1em}{.1em} 

\multicolumn{3}{c|}{Model}                                        & \multicolumn{3}{c|}{Generation Score} & \multicolumn{3}{c}{Training Resource} \\
\multicolumn{1}{c}{Name}                       & Type & \# Param\textsuperscript{$\ddagger$} & \hspace{0.2cm}FID↓\hspace{0.2cm}       & IS↑        & \hspace{0.2cm}CLIP↑\hspace{0.2cm}       & Data Size     & (Batch, \# Iter)    & A100 Days \\ 
\hline

SDM-v1.4~\cite{ldm2022,sdm_v1.4_hf}\textsuperscript{$\dagger$}     & DF   & 1.04B     & 13.05      & 36.76      & 0.2958      & >2000M\textsuperscript{$\star$}          & (2048, 1171K)  & 6250      \\ 
Small Stable Diffusion~\cite{ofasys}\textsuperscript{$\dagger$}                        & DF   & 0.76B     & 12.76      & 32.33      & 0.2851      & 229M          & (128, 1100K)   & -      \\
\rowcolor[HTML]{ECF4FF} 
BK-SDM-Base [Ours]\textsuperscript{$\dagger$}                             & DF   & 0.76B     & 15.76      & 33.79      & 0.2878      & 0.22M         & (256, 50K)   & 13        \\
\rowcolor[HTML]{ECF4FF} 
BK-SDM-Small [Ours]\textsuperscript{$\dagger$}                            & DF   & 0.66B     & 16.98      & 31.68      & 0.2677      & 0.22M         & (256, 50K)   & 13        \\
\rowcolor[HTML]{ECF4FF} 
BK-SDM-Tiny [Ours]\textsuperscript{$\dagger$}                             & DF   & 0.50B     & 17.12      & 30.09      & 0.2653      & 0.22M         & (256, 50K)   & 13        \\ 

\hline\hline
SDM-v2.1-base~\cite{ldm2022,sdm_v2-1-base_hf}\textsuperscript{$\dagger$}  & DF   & 1.26B     & 13.93      & 35.93      & 0.3075      & >2000M\textsuperscript{$\star$}       & (2048, 1620K)    & 8334     \\ \rowcolor[HTML]{ECF4FF} 
BK-SDM-v2-Base [Ours]\textsuperscript{$\dagger$}                          & DF   & 0.98B     & 15.85      & 31.70      & 0.2868      & 0.22M         & (128, 50K)   & 4       \\
\rowcolor[HTML]{ECF4FF} 
BK-SDM-v2-Small [Ours]\textsuperscript{$\dagger$}                         & DF   & 0.88B     & 16.61      & 31.73      & 0.2901      & 0.22M         & (128, 50K)   & 4        \\
\rowcolor[HTML]{ECF4FF} 
BK-SDM-v2-Tiny [Ours]\textsuperscript{$\dagger$}                          & DF   & 0.72B     & 15.68         & 31.64         & 0.2897          & 0.22M         & (128, 50K)    & 4       \\ 

\hline\hline
DALL·E~\cite{dalle1}          & AR   & 12B       & 27.5       & 17.9       & -           & 250M          & (1024, 430K)    & -     \\
CogView~\cite{cogview1}        & AR   & 4B        & 27.1       & 18.2       & -           & 30M           & (6144, 144K)    & -     \\
CogView2~\cite{cogview2}     & AR   & 6B        & 24         & 22.4       & -           & 30M           & (4096, 300K)    & -     \\
Make-A-Scene~\cite{makeScene} & AR   & 4B        & 11.84      & -          & -           & 35M           & (1024, 170K)    & -     \\
LAFITE~\cite{lafite2022}      & GAN  & 0.23B     & 26.94      & 26.02      & -           & 3M            & -      & -             \\
GALIP (CC12M)~\cite{galip}\textsuperscript{$\dagger$}     & GAN  & 0.32B     & 13.86      & 25.16          &  0.2817           & 12M            & -     & -               \\

GigaGAN~\cite{kang2023gigagan}                                       & GAN  & 1.1B      & 9.09       & -          & -           & >100M\textsuperscript{$\star$}        & (512, 1350K)     & 4783    \\
GLIDE~\cite{glide}            & DF   & 3.5B      & 12.24      & -          & -           & 250M          & (2048, 2500K)    & -    \\
LDM-KL-8-G~\cite{ldm2022}     & DF   & 1.45B     & 12.63      & 30.29      & -           & 400M          & (680, 390K)     & -     \\
DALL·E-2~\cite{dalle2}          & DF   & 5.2B      & 10.39      & -          & -           & 250M          & (4096, 3400K)     & -   \\
SnapFusion~\cite{li2023snapfusion}                                       & DF  & 0.99B      & $\sim$13.6       & -          & $\sim$0.295           & >100M\textsuperscript{$\star$}        & (2048, -)     & >128\textsuperscript{$\star$}    \\
Würstchen-v2~\cite{wuerstchen}\textsuperscript{$\dagger$}                                       & DF  & 3.1B & 22.40       & 32.87          & 0.2676           & 1700M       & (1536, 1725K)     & 1484  \\

MobileDiffusion~\cite{mobile-diffusion}\textsuperscript{$\dagger$}                                       & DF  & 0.52B & 9.01       & -         & -           & 150M       & (2048, 1000K)     & -   \\

\specialrule{.2em}{.1em}{.1em} 

\end{tabular}
\begin{tablenotes}[para,flushleft]
\footnotesize
 \textsuperscript{$\dagger$}: Evaluated with the released checkpoints.
  \textsuperscript{$\ddagger$}: Total parameters for T2I synthesis.
 \textsuperscript{$\star$}: Estimated based on public information. 
DF and AR: diffusion and autoregressive models. ↓ and ↑: lower and higher values are better.

\end{tablenotes}
\end{threeparttable}
\end{adjustbox}
\end{table*}

\subsection{Comparison with Existing Works} \label{main_results}

\noindent \textbf{Quantitative Comparison.} Tab.~\ref{table:main_comp} shows the zero-shot results for general-purpose T2I. Despite being trained with only 0.22M samples and having fewer than 1B parameters, our compressed models demonstrate competitive performance on par with existing large models.

\begin{figure*}[t!]
  \centering
    \includegraphics[width=\linewidth]{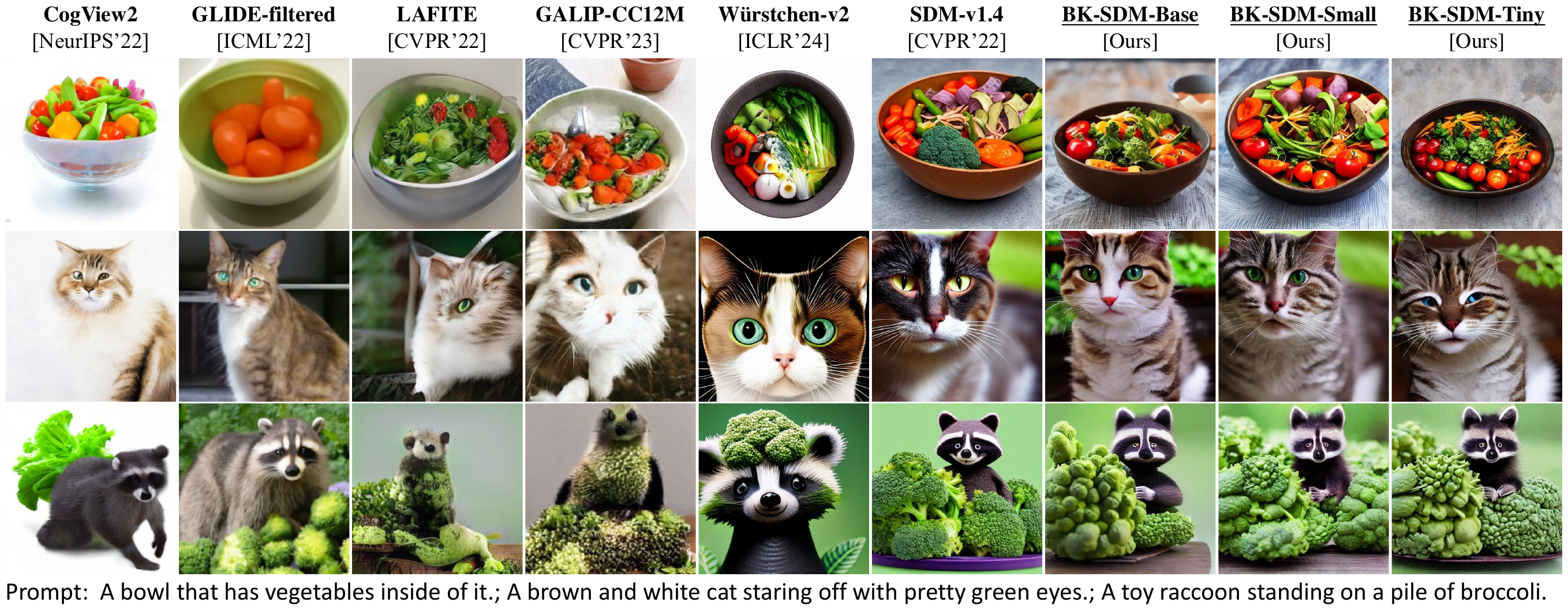}
  \caption{\textbf{Visual comparison with open-sourced models.} The results~\cite{cogview2,glide,lafite2022,galip,wuerstchen} were obtained with their official codes.}
  \label{fig:maincomp}
\end{figure*}

\noindent \textbf{Visual Comparison.} Fig.~\ref{fig:maincomp} depicts synthesized images with some MS-COCO captions. Our compact models inherit the superiority of SDM and produce more photorealistic images compared to the AR-based~\cite{cogview2} and GAN-based~\cite{lafite2022,galip} baselines. Noticeably, the same latent code results in a shared visual style between the original and our models (6th–9th columns in Fig.~\ref{fig:maincomp}), similar to the observation in transfer learning for GANs~\cite{mo2020freeze}. See Sec.~\ref{appendix:main_vis_comp} for additional results.

\begin{table*}[t!]
\centering

\caption{\textbf{Impact of compute reduction in U-Net on the entire SDM.} The number of sampling steps is indicated with the parentheses, e.g., U-Net (1) for one step. The full computation (denoted by \enquote{Whole}) covers the text encoder, U-Net, and image decoder. All corresponding values are obtained on the generation of a single 512×512 image with 25 denoising steps. The latency was measured on Xeon Silver 4210R CPU 2.40GHz and NVIDIA GeForce RTX 3090 GPU.} \label{table:compute}
\begin{adjustbox}{max width=\linewidth}
\begin{tabular}{c|cc|ccc|ccc|ccc}
\specialrule{.2em}{.1em}{.1em} 
                        & \multicolumn{2}{c|}{\# Param}                                                                                                                  & \multicolumn{3}{c|}{MACs}                                                                                                                                                                                                              & \multicolumn{3}{c|}{CPU Latency}                                                                                                                                                                                                     & \multicolumn{3}{c}{GPU Latency}                                                                                                                                                                                                    \\
\multirow{-2}{*}{Model} & U-Net                                                                              & Whole                                                      & U-Net (1)                                                                          & U-Net (25)                                                                          & Whole                                                       & U-Net (1)                                                                         & U-Net (25)                                                                         & Whole                                                       & U-Net (1)                                                                          & U-Net (25)                                                                        & Whole                                                     \\ \hline
SDM-v1.4~\cite{sdm_v1.4_hf} & 860M                                                     & 1033M                                                    & 339G                                                    & 8469G                                                  & 9716G                                                     & 5.63s                                                    & 146.28s                                                    & 153.00s                                                     & 0.049s                                                    & 1.28s                                                     & 1.41s                                                     \\ \hline
\begin{tabular}[c]{@{}c@{}}BK-SDM-\\Base~[Ours]\end{tabular}              & \cellcolor[HTML]{ECF4FF}\begin{tabular}[c]{@{}c@{}}580M\\ (-32.6\%)\end{tabular} & \begin{tabular}[c]{@{}c@{}}752M\\ (-27.1\%)\end{tabular} & \cellcolor[HTML]{ECF4FF}\begin{tabular}[c]{@{}c@{}}224G\\ (-33.9\%)\end{tabular} & \begin{tabular}[c]{@{}c@{}}5594G\\ (-33.9\%)\end{tabular} & \begin{tabular}[c]{@{}c@{}}6841G\\ (-29.5\%)\end{tabular} & \cellcolor[HTML]{ECF4FF}\begin{tabular}[c]{@{}c@{}}3.84s\\ (-31.8\%)\end{tabular} & \begin{tabular}[c]{@{}c@{}}99.95s\\ (-31.7\%)\end{tabular} & \begin{tabular}[c]{@{}c@{}}106.67s\\ (-30.3\%)\end{tabular} & \cellcolor[HTML]{ECF4FF}\begin{tabular}[c]{@{}c@{}}0.032s\\ (-34.6\%)\end{tabular} & \begin{tabular}[c]{@{}c@{}}0.83s\\ (-35.2\%)\end{tabular} & \begin{tabular}[c]{@{}c@{}}0.96s\\ (-31.9\%)\end{tabular} \\ \hline

\begin{tabular}[c]{@{}c@{}}BK-SDM-\\Small~[Ours]\end{tabular}       & \cellcolor[HTML]{ECF4FF}\begin{tabular}[c]{@{}c@{}}483M\\ (-43.9\%)\end{tabular} & \begin{tabular}[c]{@{}c@{}}655M\\ (-36.5\%)\end{tabular} & \cellcolor[HTML]{ECF4FF}\begin{tabular}[c]{@{}c@{}}218G\\ (-35.7\%)\end{tabular} & \begin{tabular}[c]{@{}c@{}}5444G\\ (-35.7\%)\end{tabular} & \begin{tabular}[c]{@{}c@{}}6690G\\ (-31.1\%)\end{tabular} & \cellcolor[HTML]{ECF4FF}\begin{tabular}[c]{@{}c@{}}3.45s\\ (-38.7\%)\end{tabular} & \begin{tabular}[c]{@{}c@{}}89.78s\\ (-38.6\%)\end{tabular} & \begin{tabular}[c]{@{}c@{}}96.50s\\ (-36.9\%)\end{tabular}  & \cellcolor[HTML]{ECF4FF}\begin{tabular}[c]{@{}c@{}}0.030s\\ (-38.7\%)\end{tabular} & \begin{tabular}[c]{@{}c@{}}0.77s\\ (-39.8\%)\end{tabular} & \begin{tabular}[c]{@{}c@{}}0.90s\\ (-36.1\%)\end{tabular} \\ 
\hline

\begin{tabular}[c]{@{}c@{}}BK-SDM-\\ Tiny~[Ours]\end{tabular}  & \cellcolor[HTML]{ECF4FF}\begin{tabular}[c]{@{}c@{}}324M\\ (-62.4\%)\end{tabular} & \begin{tabular}[c]{@{}c@{}}496M\\ (-51.9\%)\end{tabular} & \cellcolor[HTML]{ECF4FF}\begin{tabular}[c]{@{}c@{}}206G\\ (-39.5\%)\end{tabular} & \begin{tabular}[c]{@{}c@{}}5126G\\ (-39.5\%)\end{tabular} & \begin{tabular}[c]{@{}c@{}}6373G\\ (-34.4\%)\end{tabular} & \cellcolor[HTML]{ECF4FF}\begin{tabular}[c]{@{}c@{}}3.03s\\ (-46.2\%)\end{tabular} & \begin{tabular}[c]{@{}c@{}}78.77s\\ (-46.1\%)\end{tabular} & \begin{tabular}[c]{@{}c@{}}85.49s\\ (-44.1\%)\end{tabular}  & \cellcolor[HTML]{ECF4FF}\begin{tabular}[c]{@{}c@{}}0.026s\\ (-46.9\%)\end{tabular} & \begin{tabular}[c]{@{}c@{}}0.67s\\ (-47.7\%)\end{tabular} & \begin{tabular}[c]{@{}c@{}}0.80s\\ (-43.2\%)\end{tabular} \\ 
\specialrule{.2em}{.1em}{.1em} 
\end{tabular}
\end{adjustbox}
\end{table*}

\begin{table*}[t]
\centering\small
\begin{minipage}{.45\textwidth}
\centering
\caption{\textbf{Quality gains from distillation retraining.} MS-COCO 30K.} \label{table:ablation_kd_v1-v2}

\resizebox{\textwidth}{!}{
\begin{tabular}{cc|ccc|c}
\specialrule{.2em}{.1em}{.1em} 
\multicolumn{2}{c|}{BK-SDM}                         & \multicolumn{3}{c|}{Generation Score} & \# Param               \\
Type                                           & KD & \hspace{0.2cm}FID↓\hspace{0.2cm}       & IS↑        & \hspace{0.2cm}CLIP↑\hspace{0.2cm}       & U-Net                  \\ \hline
\multicolumn{1}{c|}{\multirow{2}{*}{Base}}     & \xmark & 23.57      & 23.02      & 0.2408      & \multirow{2}{*}{0.58B} \\
\multicolumn{1}{c|}{}                          & \cmark & 15.76      & 33.79      & 0.2878      &                        \\ \hline
\multicolumn{1}{c|}{\multirow{2}{*}{v2-Base}}  & \xmark & 16.76      & 25.88      & 0.2661      & \multirow{2}{*}{0.59B} \\
\multicolumn{1}{c|}{}                          & \cmark & 15.85      & 31.70      & 0.2868      &                        \\ \hline
\multicolumn{1}{c|}{\multirow{2}{*}{v2-Small}} & \xmark & 16.71      & 25.77      & 0.2655      & \multirow{2}{*}{0.49B} \\
\multicolumn{1}{c|}{}                          & \cmark & 16.61      & 31.73      & 0.2901      &                        \\ \hline
\multicolumn{1}{c|}{\multirow{2}{*}{v2-Tiny}}  & \xmark & 16.87      & 26.06      & 0.2678      & \multirow{2}{*}{0.33B} \\
\multicolumn{1}{c|}{}                          & \cmark & 15.68      & 31.64      & 0.2897      &                        \\ 

\specialrule{.2em}{.1em}{.1em} 
\end{tabular}
}
\end{minipage}
~~~
\hspace{-0.35cm}
\begin{minipage}{.53\textwidth}
\centering
\caption{\textbf{Significance of each element in transferred knowledge.} Results of BK-SDM-v2-Small on MS-COCO 30K.} \label{table:ablation_init_kd_elements}
\resizebox{\textwidth}{!}{
\begin{tabular}{c|cc|ccc}
\specialrule{.2em}{.1em}{.1em} 
\multirow{2}{*}{\begin{tabular}[c]{@{}c@{}}Weight\\Initialization\end{tabular}} & \multirow{2}{*}{\begin{tabular}[c]{@{}c@{}}Output\\KD\end{tabular}} & \multirow{2}{*}{\begin{tabular}[c]{@{}c@{}}Feature\\KD\end{tabular}} & \multicolumn{3}{c}{Generation Score} \\
                                                                       &                                                                      &                                                                       & \hspace{0.2cm}FID↓\hspace{0.2cm}       & IS↑        & \hspace{0.2cm}CLIP↑\hspace{0.2cm}      \\ \hline
Random                                                                 & \xmark                                                                   & \xmark                                                                    & 41.75      & 15.42      & 0.1733     \\
Random                                                                 & \cmark                                                                   & \xmark                                                                    & 29.01      & 18.29      & 0.1967     \\
Random                                                                 & \cmark                                                                   & \cmark                                                                    & 24.47      & 22.37      & 0.2323     \\ \hline
Teacher                                                                & \xmark                                                                   & \xmark                                                                    & 16.71      & 25.77      & 0.2655     \\
Teacher                                                                & \cmark                                                                   & \xmark                                                                    & 14.27      & 29.47      & 0.2777     \\
Teacher                                                                & \cmark                                                                   & \cmark                                                                    & 16.61      & 31.73      & 0.2901     \\ 
\specialrule{.2em}{.1em}{.1em} 
\end{tabular}
}
\end{minipage}
\end{table*}

\begin{figure*}[t!]
  \centering
   \includegraphics[width=0.88\linewidth]{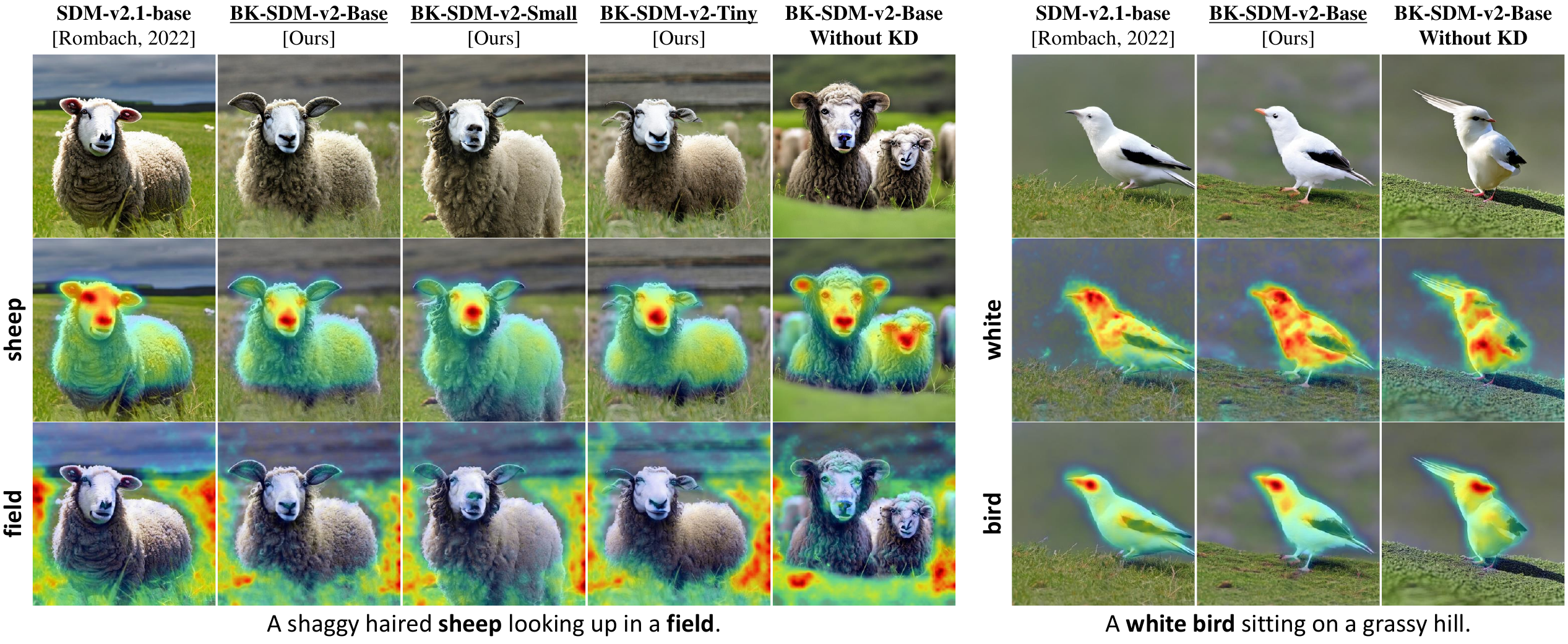}
  \caption{\textbf{Image areas affected by each word.} KD enables our models to mimic the SDM, yielding similar per-word attribution maps. The model without KD behaves differently, causing dissimilar maps and inaccurate generation (e.g., two sheep and unusual bird shapes).}
  \label{fig:attention}
\end{figure*}

\subsection{Computational Gain}  \label{sec_compute_gain}
Tab.~\ref{table:compute} shows how the compute reduction for each sampling step of the U-Net affects the overall process. The per-step reduction effectively decreases MACs and inference time by more than 30\%. Notably, BK-SDM-Tiny has 50\% fewer parameters than the original SDM.

\subsection{Benefit of Distillation Retraining} \label{sec_benefit_gain}

\noindent \textbf{T2I Performance.} Tab.~\ref{table:ablation_kd_v1-v2} summarizes the results from ablating the total KD objective (Eq. \ref{loss_out_kd}+Eq. \ref{loss_feat_kd}). Across various model types, distillation brings a clear improvement in generation quality. Tab.~\ref{table:ablation_init_kd_elements} analyzes the effect of each element in transferred knowledge. Exploiting output-level KD (Eq. \ref{loss_out_kd}) boosts the performance compared to using only the denoising task loss. Leveraging feature-level KD (Eq. \ref{loss_feat_kd}) yields further score enhancement. Additionally, using the teacher weights for initialization is highly beneficial.

\begin{table*}[t]
\centering\small
\begin{minipage}{.645\textwidth}
\centering\small
\includegraphics[width=\linewidth]{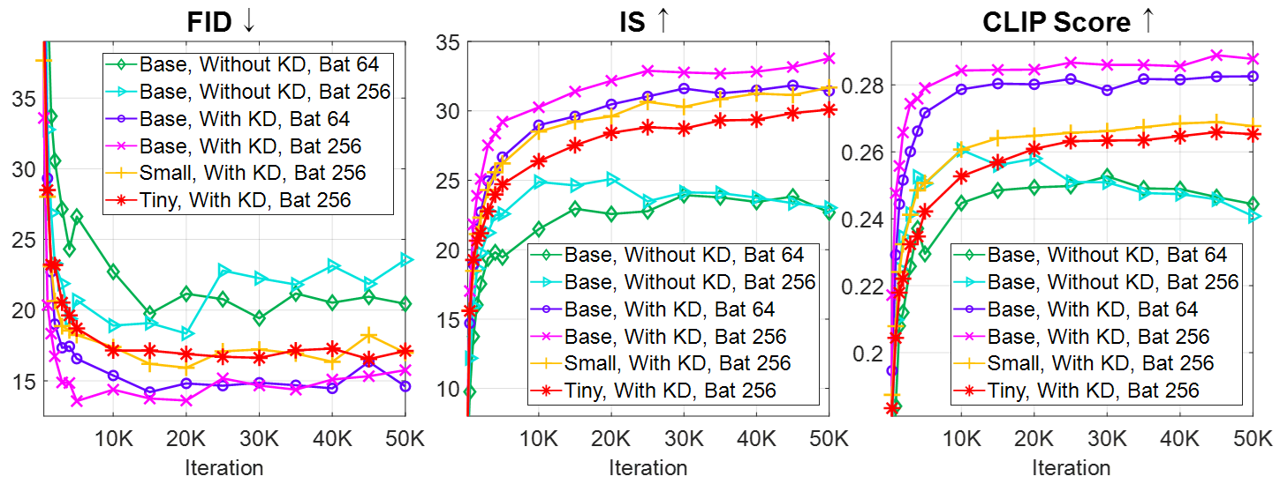}
\vspace{+0.005in}
\captionof{figure}{\textbf{Zero-shot results over training progress.} The architecture size of BK-SDM, usage of KD, and batch size are denoted. Results on MS-COCO 30K.} \label{fig:iter_scores}
\end{minipage}
~~~
\hspace{-0.3cm}
\begin{minipage}{.329\textwidth}
\centering\small
\includegraphics[width=\textwidth]{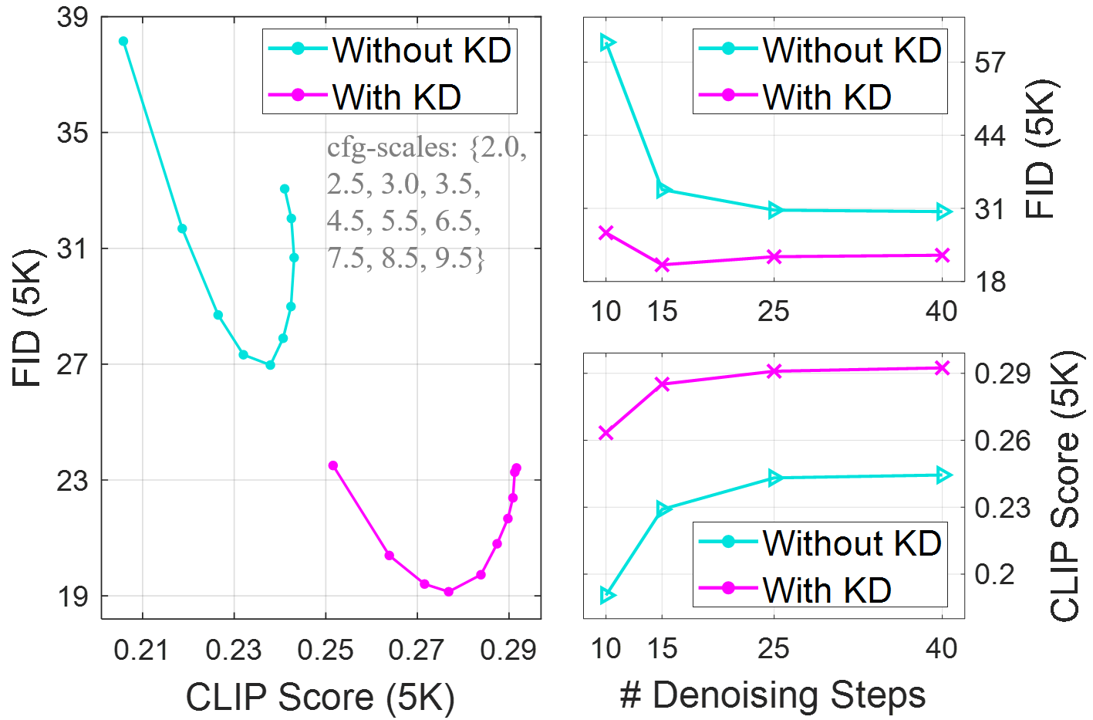}
\vspace{+0.005in}
\captionof{figure}{\textbf{Trade-off curves.} Left: FID \textit{vs.} CLIP score; Right: quality \textit{vs.} efficiency. Ours-Base on MS-COCO 5K.} \label{fig:tradeoff}
\end{minipage}
\end{table*}

\noindent \textbf{Cross-attention Resemblance.} Fig.~\ref{fig:attention} displays the per-word attribution maps~\cite{tang2023daam} created by aggregating cross-attention scores over spatiotemporal dimensions. The attribution maps of our models are semantically and spatially similar to those of the original model, indicating the merit of supervisory signals at multiple stages via KD. In contrast, the baseline model without KD activates incorrect areas, leading to text-mismatched generation results.

\noindent \textbf{Training Progress.} Fig.~\ref{fig:iter_scores} shows the results over training iterations. Without KD, training solely with the denoising task loss causes fluctuations or sudden drops in performance (indicated with green and cyan). In contrast, distillation (purple and pink) stabilizes and accelerates the training process, demonstrating the benefit of providing sufficient hints for training guidance. Notably, our small- and tiny-size models trained with KD (yellow and red) outperform the bigger base-size model without KD (cyan). Additionally, while the best FID score is observed early on, IS and CLIP score exhibit ongoing improvement, implying that judging models merely with FID may be suboptimal.

\noindent \textbf{Trade-off Results.} Fig.~\ref{fig:tradeoff} shows the results of BK-SDM-Base with and without KD on MS-COCO 512×512 5K. Higher classifier-free guidance scales~\cite{clsfreeguide,imagen} lead to better text-aligned images at the cost of less diversity. More denoising steps improve generation quality at the cost of slower inference. Distillation retraining achieves much better trade-off curves than the baseline without KD.
 
\subsection{Comparison with Different Pruning Criteria} \label{sec_prune_crit}

We extend our baseline comparisons by calculating several block-level importance scores~\cite{kim2024shortened}.\footnote{Let the $(h_{\text{out}}, h_{\text{in}})$-size weight matrix in the $b$-th block be $\mathbf{W}^{l,b} = \left[W_{i,j}^{l,b}\right]$, where $l$ denotes the layer type (\textit{e.g.}, convolution (flattened) or attention's key projection). The scores at the output neuron level~\cite{wanda} are aggregated for the block-level importance criteria, $S_{\text{Magnitude}}^b = \mathbb{E}_{l,i}\left[ \sum_j \left| W_{i,j}^{l,b} \right| \right]$ and $S_{\text{Taylor}}^b = \mathbb{E}_{l,i}\left[ \sum_j \left| \frac{\partial \mathcal{L}(D)}{\partial W_{i,j}^{l,b}} W_{i,j}^{l,b} \right| \right]$. Here, $\mathcal{L}$ and $D$ denote the denoising task loss and a calibration set of 1K samples. The final scores are then ranked to remove unimportant blocks, or to replace them with interpolation for unremovable blocks.} Furthermore, the impact of applying KD at the end of each stage is also examined. Tab.~\ref{table:prune_crit} shows the results. Magnitude (Mag) pruning\cite{li2016pruning} removes critical outer blocks, causing irrecoverable loss. Taylor pruning~\cite{lecun1989optimal,molchanov2019importance,fang2023structural} removes only the inner part, causing ineffective reduction in MACs and speed; additionally, this method demands the intensive calculation of backward gradients. The CLIP Score method  (based on Fig.~\ref{fig:prune_sensitivity_anal}) results in severe pruning of attention blocks, adversely affecting performance. Our approach achieves superior or comparable results against the benchmark methods and can be directly applied to all the SDM versions in v1 and v2. This advantage comes without the extra phase of calculating pruning criteria while ensuring a favorable balance between performance and inference speed. Notably, KD always remains effective.

\subsection{Human Preference Assessment} \label{sec_user_study}

\begin{table*}[t]
\centering\small
\begin{minipage}{.54\textwidth}
\centering
\caption{\textbf{Different block-level criteria at similar parameters.} Taylor pruning removes solely the inner blocks, leading to insufficient reduction in MACs. Our method attains a favorable compromise between performance and inference speed. Results on MS-COCO 30K.} \label{table:prune_crit}
\resizebox{\textwidth}{!}{
\begin{threeparttable}
\begin{tabular}{c|cc|cc|cc}
\specialrule{.2em}{.1em}{.1em} 
\multirow{2}{*}{\begin{tabular}[c]{@{}c@{}}Pruning\\ Criterion\end{tabular}} & \multicolumn{2}{c|}{FID↓} & \multicolumn{2}{c|}{\hspace{0.15cm}CLIP Score↑\hspace{0.15cm}} & \multicolumn{2}{c}{U-Net}           \\
                                                                             & \hspace{0.15cm}KD\xmark\hspace{0.15cm}        & \hspace{0.15cm}KD\cmark\hspace{0.15cm}        & \hspace{0.15cm}KD\xmark\hspace{0.15cm}         & \hspace{0.15cm}KD\cmark\hspace{0.15cm}        & \hspace{0.15cm}MACs\hspace{0.15cm}   & \hspace{0.15cm}\# Param\hspace{0.15cm} \\ \hline
Mag~\cite{li2016pruning}                                                                   & >150       & >150       & <0.1       & <0.1      & 122.2G & 495.5M                     \\
Taylor~\cite{lecun1989optimal,fang2023structural}                                                                        & 18.85        & 16.73         & 0.2429          & 0.2749         & 305.4G & 493.4M                     \\
CLIP [Fig.~\ref{fig:prune_sensitivity_anal}]                                                                   & 32.93         & 21.86         & 0.1936         & 0.2283         & 204.8G & 496.8M                     \\
Ours-Small                                                                         & 22.97         & 16.83         & 0.2287          & 0.2668         & 217.7G & 482.3M                     \\

\specialrule{.2em}{.1em}{.1em} 
\end{tabular}
\begin{tablenotes}[para,flushleft]
\footnotesize
Retraining with batch 128, 0.22M data, and 50K iters. 
\newline
Original SD-v1.4's U-Net: (MACs, Params) = (338.7G, 859.5M).
\end{tablenotes}
\end{threeparttable}
}
\end{minipage}
~~~
\hspace{-0.35cm}
\begin{minipage}{.44\textwidth}
\centering
\caption{\textbf{A/B-testing preference study.} Total 2,501 comparisons.} \label{rbt_table:user_study}
\resizebox{\textwidth}{!}{
\begin{threeparttable}
\begin{tabular}{c|cccc}
\specialrule{.2em}{.1em}{.1em} 
\begin{tabular}[c]{@{}c@{}}Ours-\\ Small\end{tabular} & \begin{tabular}[c]{@{}c@{}}GLIDE-filt\\ \cite{glide} \end{tabular} & \begin{tabular}[c]{@{}c@{}}GALIP\\ \cite{galip} \end{tabular}  & \begin{tabular}[c]{@{}c@{}}Würstchen-v2\\ \cite{wuerstchen} \end{tabular} & \begin{tabular}[c]{@{}c@{}}SDM-v1.4\\ \cite{sdm_v1.4_hf} \end{tabular} \\ \hline
\# Win\textsuperscript{$\dagger$}                                                & \begin{tabular}[c]{@{}c@{}}453\\ (71.5\%)\end{tabular}     & \begin{tabular}[c]{@{}c@{}}379\\ (61.5\%)\end{tabular} & \begin{tabular}[c]{@{}c@{}}319\\ (51.4\%)\end{tabular}       & \begin{tabular}[c]{@{}c@{}}202\\ (32.1\%)\end{tabular}   \\ \hline
\# Defeat\textsuperscript{$\dagger$}                                             & \begin{tabular}[c]{@{}c@{}}181\\ (28.5\%)\end{tabular}     & \begin{tabular}[c]{@{}c@{}}237\\ (38.5\%)\end{tabular} & \begin{tabular}[c]{@{}c@{}}302\\ (48.6\%)\end{tabular}       & \begin{tabular}[c]{@{}c@{}}428\\ (67.9\%)\end{tabular}   \\ 
\specialrule{.2em}{.1em}{.1em} 
\end{tabular}
\begin{tablenotes}[para,flushleft]
\footnotesize
\textsuperscript{$\dagger$}Participants selected preferred images from pairs with the same prompts. One image in each pair was always from BK-SDM-Small, while the other was randomly chosen from the four methods.
\end{tablenotes}
\end{threeparttable}
}
\end{minipage}
\end{table*}

\begin{table*}[t]
\centering\small
\begin{minipage}{.55\textwidth}
\centering
\caption{\textbf{Impact of training batch size and iterations.} Results on MS-COCO 30K.} \label{table:ablation_batch_iter}
\resizebox{\textwidth}{!}{
\begin{tabular}{c|cc|cc|cc}
\specialrule{.2em}{.1em}{.1em} 

BK-SDM      & \multicolumn{2}{c|}{Base} & \multicolumn{2}{c|}{Small} & \multicolumn{2}{c}{Tiny} \\ \hline
Batch Size  & 64          & 256         & 64           & 256         & 64          & 256        \\ \hline
FID↓        & 14.61       & 15.76       & 16.87        & 16.98       & 17.28       & 17.12      \\
IS↑         & 31.44       & 33.79       & 29.51        & 31.68       & 28.33       & 30.09      \\
CLIP↑ & 0.2826      & 0.2878      & 0.2644       & 0.2677      & 0.2607      & 0.2653     \\ 
\specialrule{.2em}{.1em}{.1em} 

\specialrule{.2em}{.1em}{.1em} 

BK-SDM  & \multicolumn{2}{c|}{Base (Data 2.3M)} & \multicolumn{2}{c|}{Small (Data 2.3M)} & \multicolumn{2}{c}{Tiny (Data 2.3M)} \\ \hline
\# Iter & 50K               & 100K              & 50K                & 100K              & 50K               & 100K             \\ \hline
FID↓    & 14.81             & 15.39             & 17.05              & 17.01             & 17.53             & 17.63            \\
IS↑     & 34.17             & 34.76             & 33.10              & 33.14             & 31.32             & 32.26            \\
CLIP↑   & 0.2883            & 0.2889            & 0.2734             & 0.2754            & 0.2690            & 0.2713           \\ 

\specialrule{.2em}{.1em}{.1em} 
\end{tabular}

}
\end{minipage}
~~~
\hspace{-0.35cm}
\begin{minipage}{.431\textwidth}
\centering
\includegraphics[width=\textwidth]{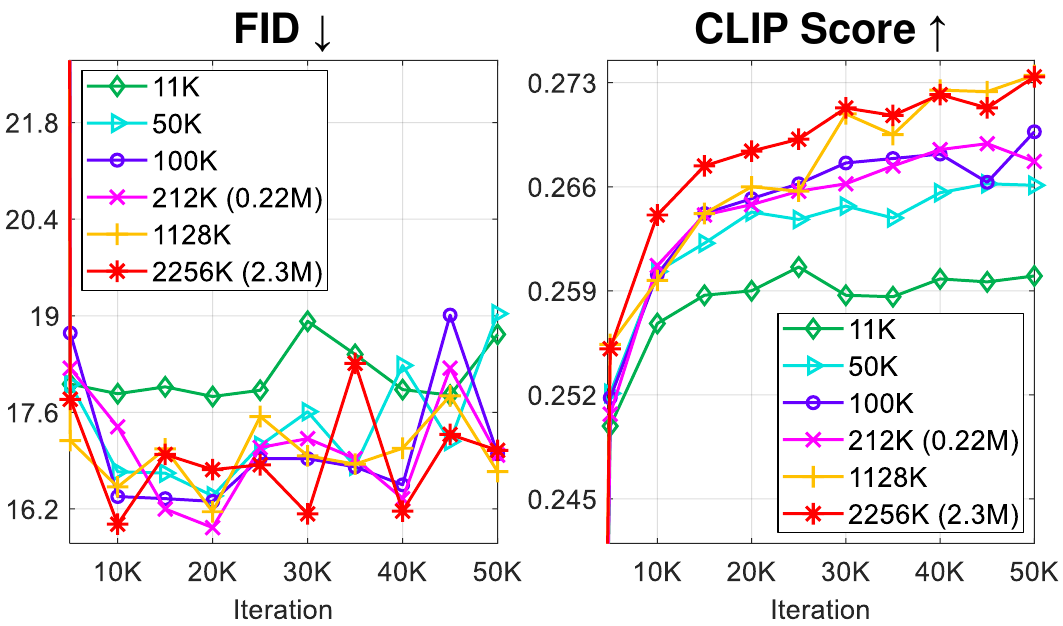}
\vspace{+0.01in}
\captionof{figure}{\textbf{Impact of training data volume.} Results of BK-SDM-Small on MS-COCO 30K.} \label{fig:datasize_iter}
\end{minipage}
\end{table*}

Despite their extensive use, automated metrics for assessing image quality (e.g., FID) do not consistently match with human evaluation~\cite{imagen,dalle2}. To further address this, we have conducted an A/B testing-based study: participants were shown pairs of images generated using the same MS-COCO prompt and asked to choose their preferred image in relation to the prompt. One image in each pair was always from our model, while the other was randomly selected from one of four comparative methods. A total of 25 participants conducted 2,501 comparisons online. The participants had no bias towards the examined models and were not compensated. The prompts were randomly drawn from the MS-COCO set and varied for 3 to 5 participants. As shown in Tab.~\ref{rbt_table:user_study}, our model (0.66B parameters) excels over GLIDE-filtered, GALIP, and Würstchen-v2 (3.1B).

\begin{table*}[t!]
\centering\small
\begin{minipage}{.67\textwidth}
\centering
\caption{\textbf{Personalized generation with finetuning over different backbones.} Our compact models can preserve subject fidelity (DINO and CLIP-I) and prompt fidelity (CLIP-T) of the original SDM with reduced finetuning (FT) costs and fewer parameters.} \label{table:dreambooth}
\resizebox{\textwidth}{!}{
\begin{threeparttable}
\begin{tabular}{l|ccc|c}
\specialrule{.2em}{.1em}{.1em} 
\multicolumn{1}{c|}{Backbone \{Param\}} & DINO↑ & CLIP-I↑ & CLIP-T↑ & FT (Time, Mem)\textsuperscript{$\dagger$}                         \\ \hline
SDM-v1.4~\cite{sdm_v1.4_hf} \{1.04B\}            & 0.728 & 0.725   & 0.263   & (882s, 23.0GB)                         \\ \hline
\rowcolor[HTML]{ECF4FF} 
BK-SDM-Base \{0.76B\}                   & 0.723 & 0.717   & 0.260   & (623s, 18.7GB)                         \\
\rowcolor[HTML]{ECF4FF} 
BK-SDM-Small \{0.66B\}                  & 0.720 & 0.705   & 0.259   & \cellcolor[HTML]{ECF4FF}(604s, 17.2GB) \\
\rowcolor[HTML]{ECF4FF} 
BK-SDM-Tiny \{0.50B\}                   & 0.715 & 0.693   & 0.261   & (560s, 13.1GB)                         \\ \hline
Base (Batch 64) \{0.76B\}                       & 0.718 & 0.708   & 0.262   & (623s, 18.7GB)                         \\
- No KD \& Random Init.               & 0.594 & 0.465   & 0.191   & (623s, 18.7GB)                         \\
- No KD \& Teacher Init.              & 0.716 & 0.669   & 0.258   & (623s, 18.7GB)                         \\ 
\specialrule{.2em}{.1em}{.1em} 
\end{tabular}

\begin{tablenotes}[para,flushleft]
\footnotesize
\textsuperscript{$\dagger$}: Per-subject FT time and GPU memory for 800 iterations on RTX 3090.
\end{tablenotes}
\end{threeparttable}
}
\end{minipage}
~~~
\hspace{-0.35cm}
\begin{minipage}{.31\textwidth}
\centering
\caption{\textbf{Speedups on edge devices.}} \label{table:edge_time}
\resizebox{\textwidth}{!}{
\begin{tabular}{c|cc}
\specialrule{.2em}{.1em}{.1em} 

Model                                                   & \begin{tabular}[c]{@{}c@{}}AGX Orin\\ 32GB\end{tabular} & \hspace{0.15cm}iPhone 14                                              \\ \hline
SDM-v1                                                  & 4.9s~\cite{sdm_v1.5_hf}                                         & 5.6s~\cite{sdm_v1.4_hf}                                        \\ \hline
\begin{tabular}[c]{@{}c@{}}BK-SDM-\\ Base\end{tabular}  & \begin{tabular}[c]{@{}c@{}}3.4s\\ (-31\%)\end{tabular}  & \begin{tabular}[c]{@{}c@{}}4.0s\\ (-29\%)\end{tabular} \\ \hline
\begin{tabular}[c]{@{}c@{}}BK-SDM-\\ Small\end{tabular} & \begin{tabular}[c]{@{}c@{}}3.2s\\ (-35\%)\end{tabular}  & \begin{tabular}[c]{@{}c@{}}3.9s\\ (-29\%)\end{tabular} \\ \hline
\begin{tabular}[c]{@{}c@{}}BK-SDM-\\ Tiny\end{tabular}  & \begin{tabular}[c]{@{}c@{}}2.8s\\ (-43\%)\end{tabular}  & \begin{tabular}[c]{@{}c@{}}3.9s\\ (-29\%)\end{tabular} \\ 

\specialrule{.2em}{.1em}{.1em} 
\end{tabular}
}
\end{minipage}
\end{table*}

\begin{figure}[t!]
  \centering
    \includegraphics[width=\linewidth]{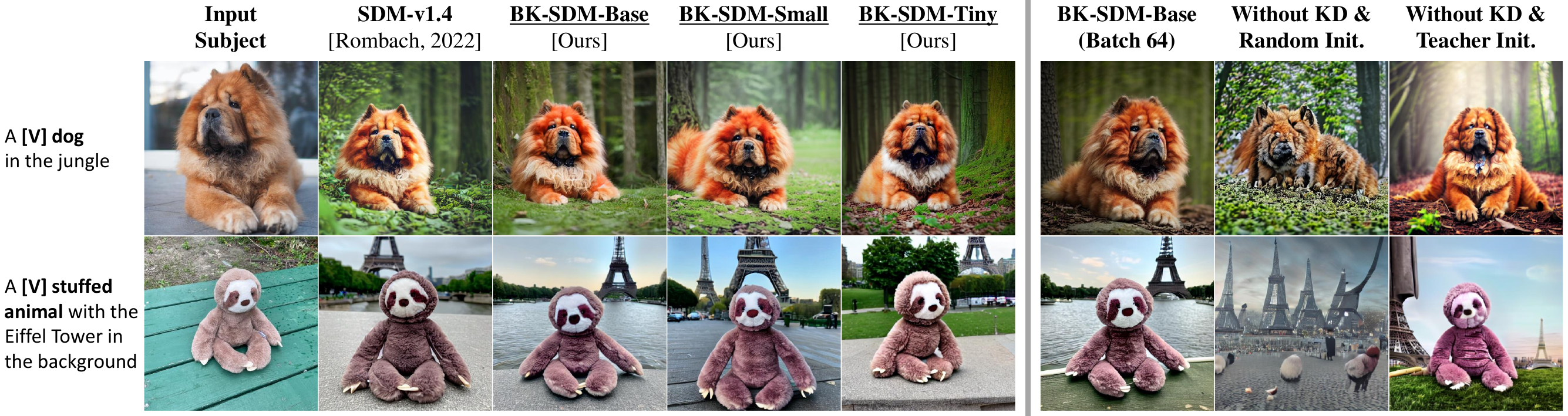}   
  \caption{\textbf{Visual results of personalized generation.} Each subject is marked as ``a [identifier] [class noun]” (e.g., ``a [V] dog").} \label{fig:dreambooth}

\end{figure}

\begin{table*}[t!]
\centering
\begin{minipage}{.665\linewidth}
\centering\small
\includegraphics[width=\textwidth]{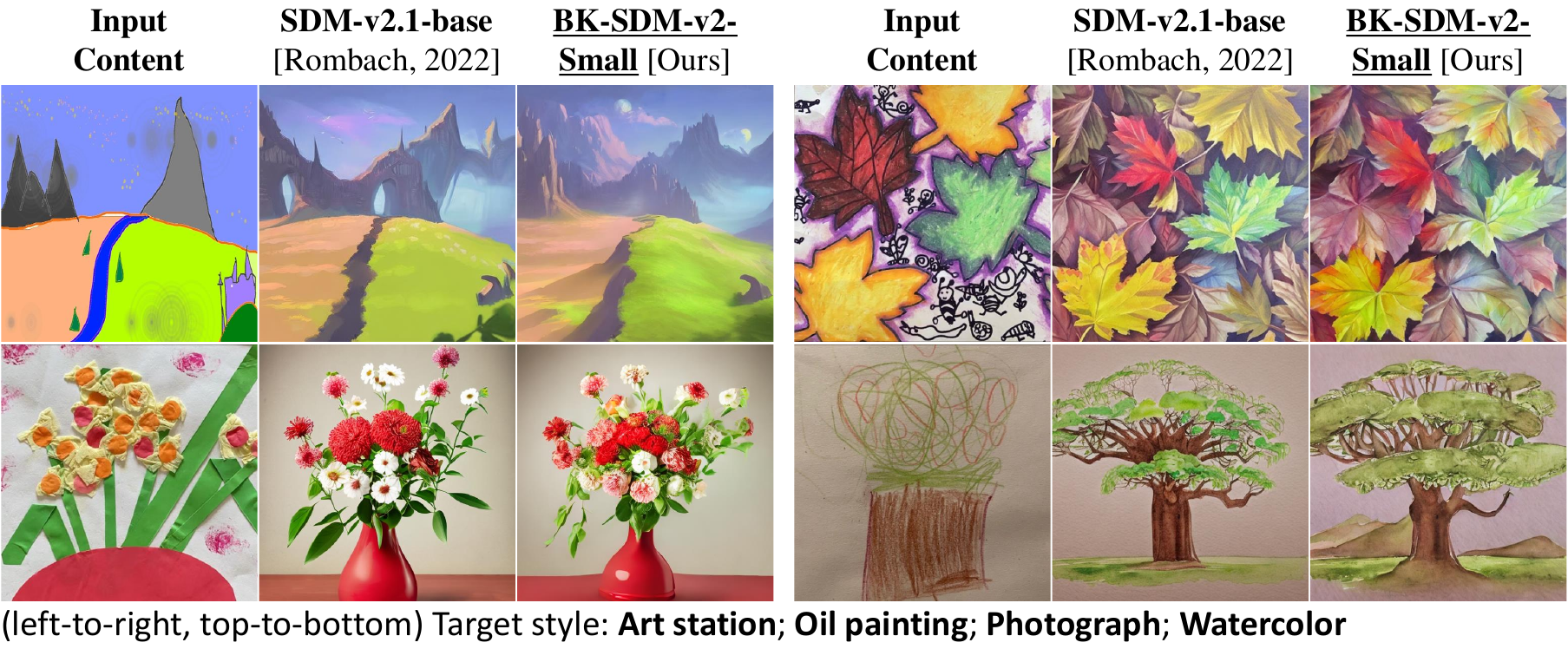}
\vspace{+0.005in}
\captionof{figure}{\textbf{Text-guided image-to-image translation.}} \label{fig:i2i}
\end{minipage}
~~~
\hspace{-0.36cm}
\begin{minipage}{.31\linewidth}
\centering\small
\includegraphics[width=\textwidth]{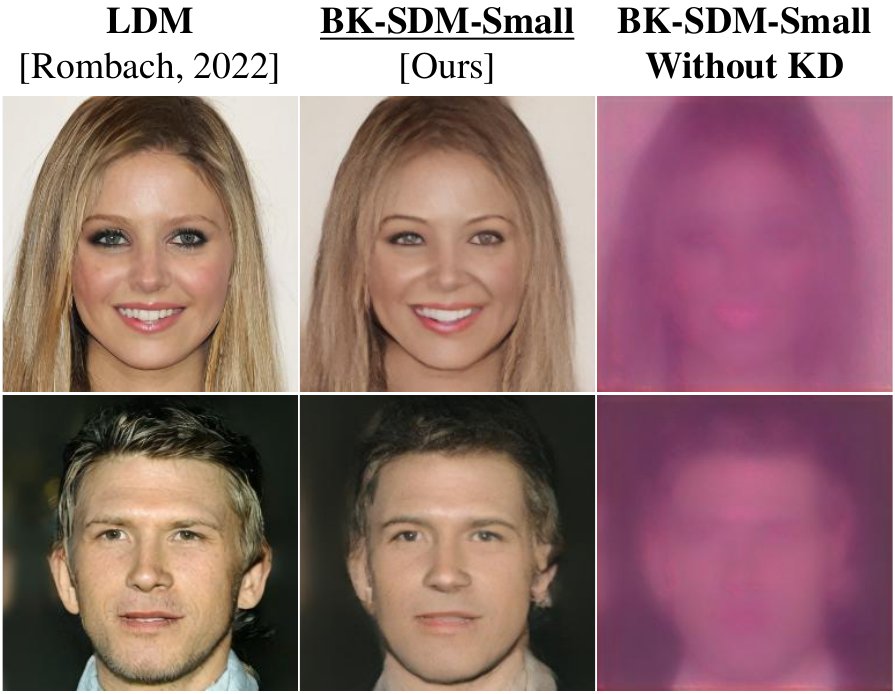}
\vspace{+0.005in}
\captionof{figure}{\textbf{Small LDM for face synthesis.}} \label{fig:ldm_face}
\end{minipage}
\end{table*}

\subsection{Impact of Training Resources on Performance} \label{sec_train_resources}
Consistent with existing evidence, the use of larger batch sizes, more extensive data, and longer iterations for training enhances performance in our work (see Tab.~\ref{table:ablation_batch_iter}, Fig.~\ref{fig:datasize_iter}, and Sec.~\ref{appendix:data_volume}). However, this benefit requires increased resource demands (e.g., extended training days without multiple high-spec GPUs and greater data storage capacity). As such, despite the better performing models in Tab.~\ref{table:ablation_batch_iter}, we primarily report the models with fewer resources. We believe that accessible training costs by many researchers can help drive advancement in massive models.

\subsection{Application}

\noindent \textbf{Personalized T2I Synthesis.} Tab.~\ref{table:dreambooth} compares the fine-tuning results with DreamBooth~\cite{dreambooth} over different backbones to create images about a given subject. BK-SDMs can preserve 95\%$\sim$99\% scores of the original SDM while cutting fine-tuning costs. Fig.~\ref{fig:dreambooth} depicts that our models can accurately capture the subject details and generate various scenes. Over the models retrained with a batch size of 64, the baselines without KD fail to generate the subjects or cannot maintain the identity details. See Sec.~\ref{appendix_dreambooth} for further results.

\noindent \textbf{Image-to-Image Translation.} Fig.~\ref{fig:i2i} presents the text-guided stylization results with SDEdit~\cite{sdedit}. Our model, resembling the ability of the original SDM, faithfully produces images given style-specified prompts and content pictures. See Sec.~\ref{appendix:img2img} for additional results.

\noindent \textbf{Deployment on Edge Devices.} We deploy our models trained with 2.3M pairs and compare them against the original SDM under the same setup on edge devices (20 denoising steps on NVIDIA Jetson AGX Orin 32GB and 10 steps on iPhone 14). Our models produce a 512×512 image within 4 seconds (see Tab.~\ref{table:edge_time}), while maintaining acceptable image quality (Fig.~\ref{fig:teaser}(d) and Sec.~\ref{appendix_edge_deploy}).

\noindent \textbf{Another LDM.} SDMs are derived from LDMs~\cite{ldm2022}, which share a similar U-Net design across many tasks. To validate the generality of our work, we compress an LDM (with 308M parameters and 410K training iterations) for unconditional generation on CelebA-HQ~\cite{ldm-face} by using the same approach of BK-SDM-Small (187M parameters and 30K iterations). Fig.~\ref{fig:ldm_face} shows the efficacy of our architecture and distillation retraining.

\section{Conclusion and Discussion} \label{sec:conclusion}

We uncover the potential of architectural compression for general-purpose text-to-image synthesis with a renowned model, Stable Diffusion. Our block-removed lightweight models are effective for zero-shot generation, achieving competitive results against large-scale baselines. Distillation is a key of our method, leading to powerful retraining even under very constrained resources. Our work is orthogonal to previous works for efficient diffusion models, e.g., enabling fewer sampling steps, and can be readily combined with them. We hope our study can facilitate future research on structural compression of large diffusion models.

\noindent\textbf{Limitations.} Our compact models inherit the capability of the source model for high-fidelity image generation, but they have shortcomings such as inaccurate generation of full-body human appearance. 

\noindent\textbf{Negative Social Impacts.} Recent large generative models are capable of creating high-quality plausible content. To avoid causing unintended social usage, researchers should take steps to ensure the appropriateness of the training data.

\section*{Acknowledgments}
We thank the Microsoft Startups Founders Hub program and the AI Industrial Convergence Cluster Development project funded by the Ministry of Science and ICT (MSIT, Korea) and Gwangju Metropolitan City for their generous support of GPU resources.

\bibliographystyle{splncs04}
\bibliography{main}

\clearpage
\setcounter{page}{1}
\appendix
\onecolumn

\begin{center}

{\bf {\Large Appendix of BK-SDM}} 
\end{center}

\section{U-Net Architecture and Distillation Retraining}\label{appendix:model_details}

\begin{figure*}[h]
  \centering
    \includegraphics[width=\linewidth]{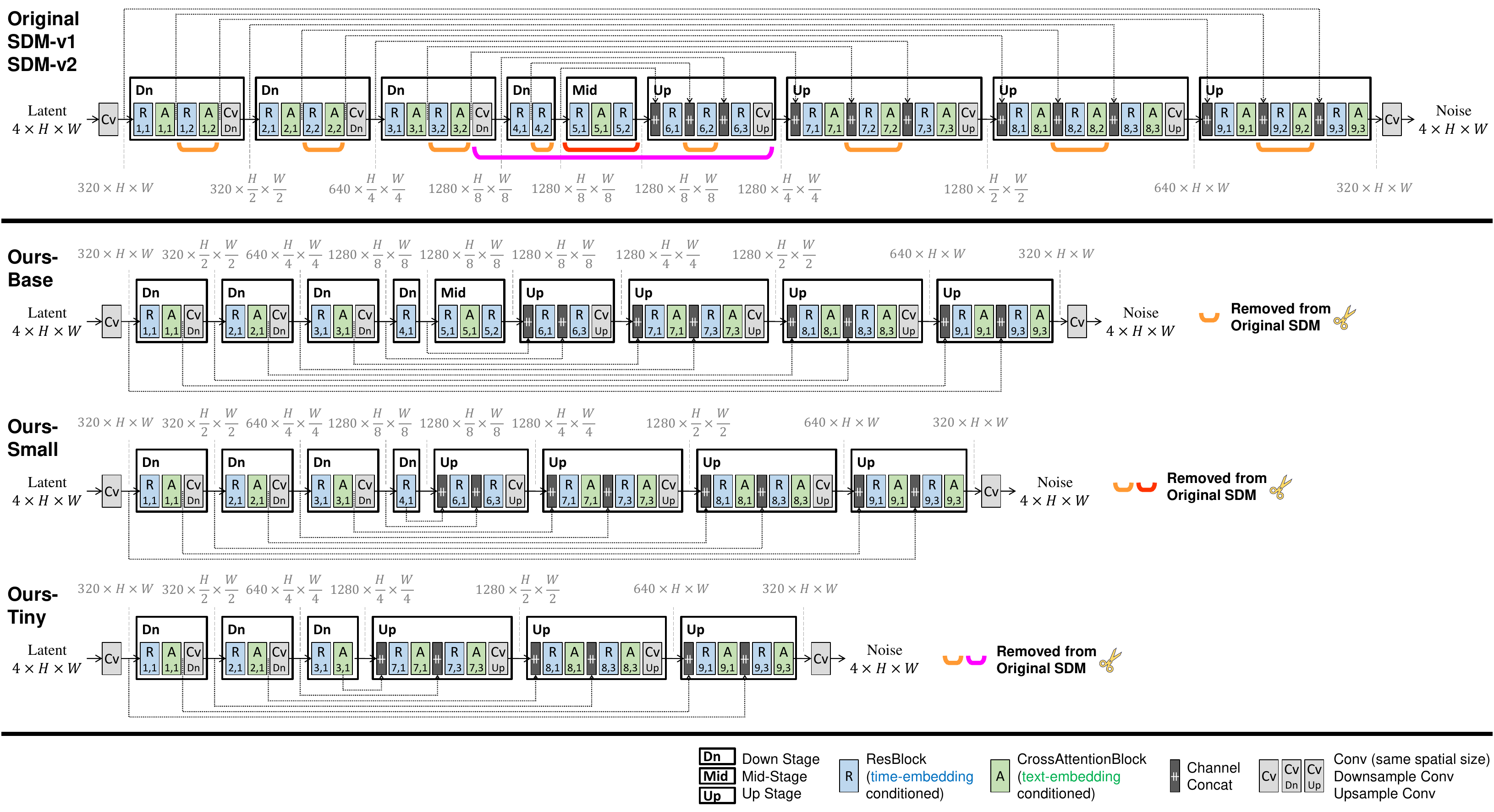}
  \caption{U-Net architectures of SDM-v1, SDM-v2, and BK-SDMs.}
  \label{fig_supple_arch}
\end{figure*}

\begin{figure*}[h!]
  \centering
    \includegraphics[width=\linewidth]{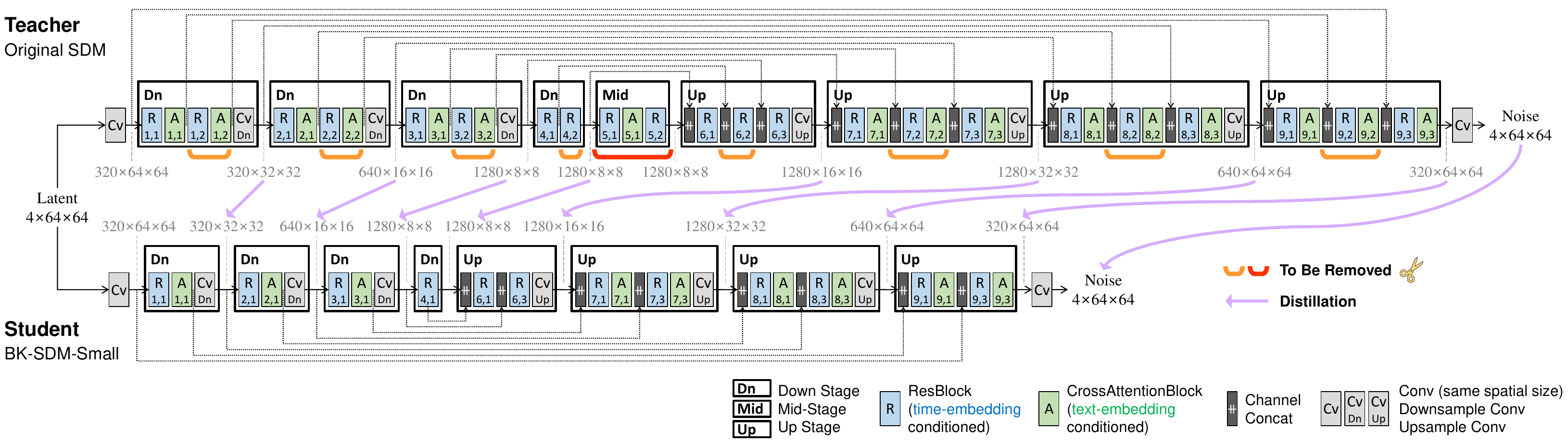}
  \caption{Distillation retraining process. The compact U-Net student is built by eliminating several residual and attention blocks from the original U-Net teacher. Through the feature and output distillation from the teacher, the student can be trained effectively yet rapidly. The default latent resolution for SDM-v1 and v2-base is $H=W=64$ in Fig.~\ref{fig_supple_arch}, resulting in 512×512 generated images.}
  \label{fig_supple_kd}
\end{figure*}

\begin{figure*}[h]
  \centering
    \includegraphics[width=\linewidth]{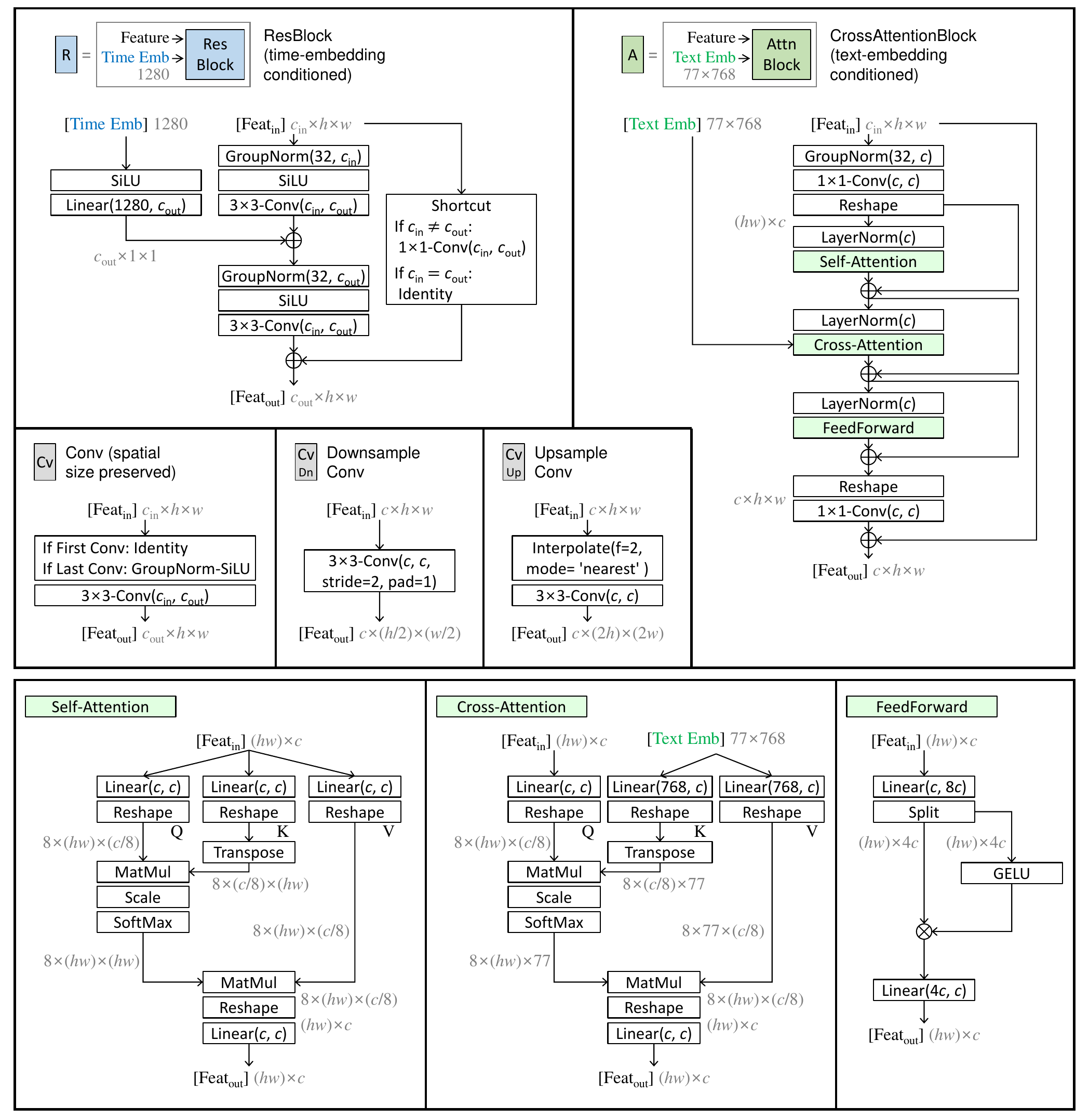}
  \caption{Block components in the U-Net.}
  \label{fig_supple_block}
\end{figure*}

Figs.~\ref{fig_supple_arch} and \ref{fig_supple_kd} depict the U-Net architectures and distillation process, respectively. Our approach is directly applicable to all the SDM versions in v1 and v2 (i.e., v1.1/2/3/4/5, v2.0/1, and v2.0/1-base), which share the same U-Net block configuration. See Fig.~\ref{fig_supple_block} for the block details.

\clearpage

Fig.~\ref{fig_supple_block} shows the details of architectural blocks. Each residual block (ResBlock) contains two 3-by-3 convolutional layers and is conditioned on the time-step embedding. Each attention block (AttnBlock) contains a self-attention module, a cross-attention module, and a feed-forward network. The text embedding is merged via the cross-attention module. Within the attention block, the feature spatial dimensions $h$ and $w$ are flattened into a sequence length of $hw$. The number of channels $c$ is considered as an embedding size, processed with attention heads. The number of groups for the group normalization is set to 32. The differences between SDM-v1 and SDM-v2 include the number of attention heads (8 for all the stages of SDM-v1 and [5, 10, 20, 20] for different stages of SDM-v2) and the text embedding dimensions (77×768 for SDM-v1 and 77×1024 for SDM-v2). 

\clearpage

\section{Impact of Mid-stage Removal}\label{appendix:mid_removal}
\begin{figure*}[h]
  \centering
    \includegraphics[width=\linewidth]{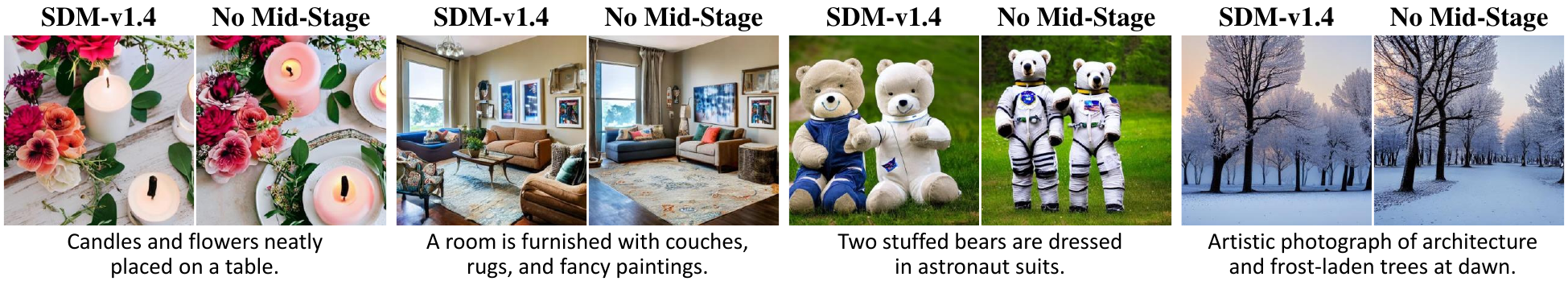}
\caption{Visual results of the mid-stage removed U-Net from SDM-v1.4~\cite{sdm_v1.4_hf}.} \label{fig_mid_rm_supple}
\end{figure*}

Removing the entire mid-stage from the original U-Net does not noticeably degrade the generation quality for many text prompts while effectively reducing the number of parameters. See Fig.~\ref{fig_mid_rm_supple} and Tab.~\ref{table:supple_mid_rm}. Retraining is not performed.

\begin{table}[h]
\centering
\caption{Minor impact of eliminating the mid-stage on MS-COCO 256×256 30K.}\label{table:supple_mid_rm}
\begin{adjustbox}{max width=0.8\linewidth}
\begin{tabular}{c|cc|cc}
\specialrule{.2em}{.1em}{.1em} 
\multirow{2}{*}{Model}                                      & \multicolumn{2}{c|}{Performance} & \multicolumn{2}{c}{\# Parameters}                                                                                          \\
                                                            & FID ↓           & IS ↑           & U-Net                                                      & Whole                                                     \\ \hline
SDM-v1.4 \cite{sdm_v1.4_hf}                                       & 13.05           & 36.76          & 859.5M                                                     & 1032.1M                                                   \\ \hline
Mid-Stage Removal & 15.60           & 32.33          & 762.5M (-11.3\%) & 935.1M (-9.4\%) \\ 
\specialrule{.2em}{.1em}{.1em} 
\end{tabular}
\end{adjustbox}
\end{table}

\clearpage
\section{Block-level Pruning Sensitivity Analysis} \label{appendix:prune_sens_anal}
\begin{figure*}[h!]
  \centering
    \includegraphics[width=\linewidth]{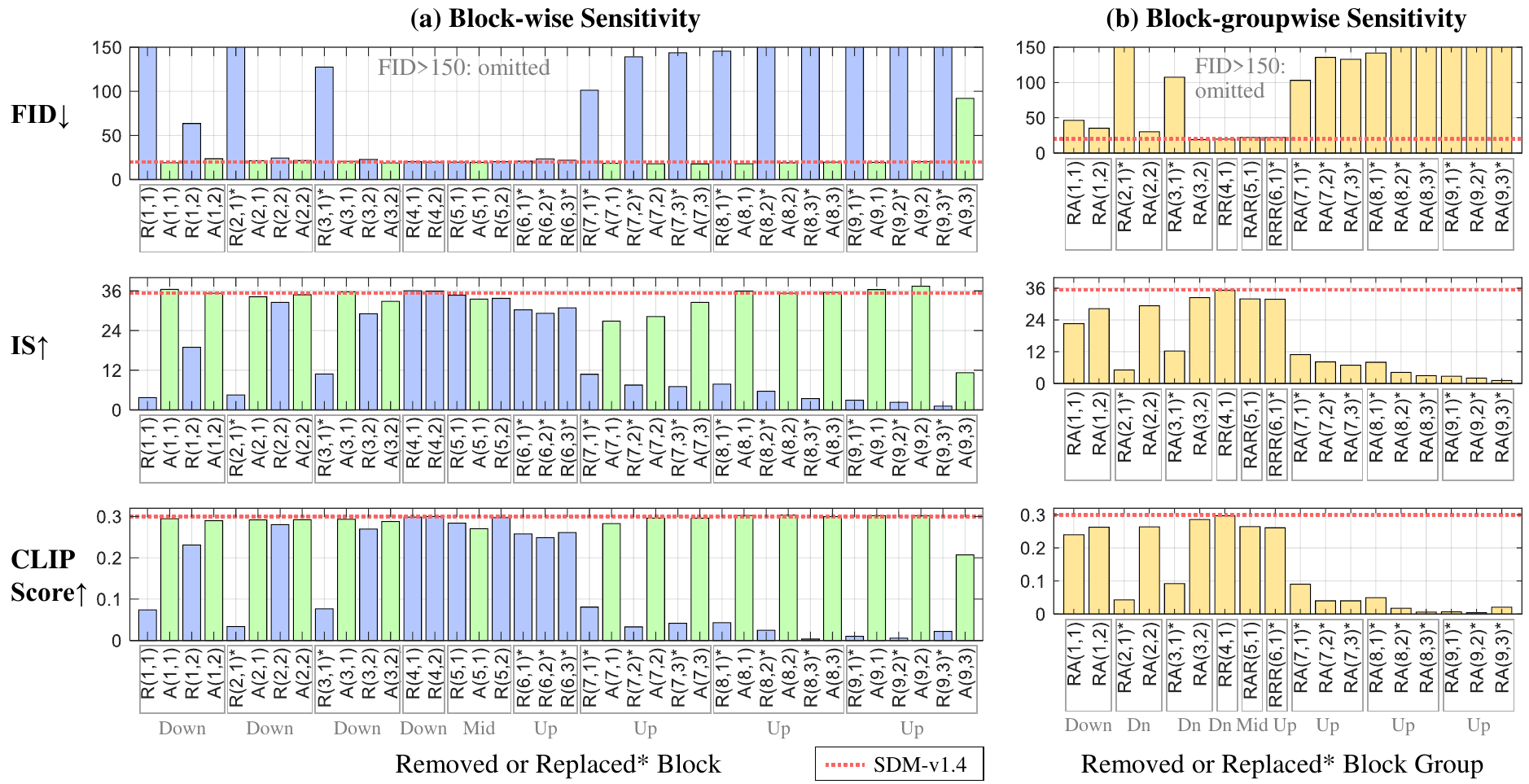}
  \caption{Analyzing the importance of (a) each block and (b) each group of paired/triplet blocks in SDM-v1.4. Evaluation on MS-COCO 512×512 5K. The block notations match Fig.~\ref{fig_supple_arch}. Whenever possible (i.e., with the same dimensions of input and output), we remove each block to examine its effect on generation performance. For blocks with different channel dimensions of input and output, we replace them with channel interpolation modules (denoted by “*”) to mimic the removal while retaining the information. The results are aligned with our architectural choices (e.g., removal of innermost stages and the second R-A pairs in down stages).} \label{fig_supple_prune_sensitivity_anal}
\end{figure*}

\clearpage

\section{Comparison with Existing Studies}\label{appendix:main_vis_comp}

\begin{figure*}[h!]
  \centering
    \includegraphics[width=\linewidth]{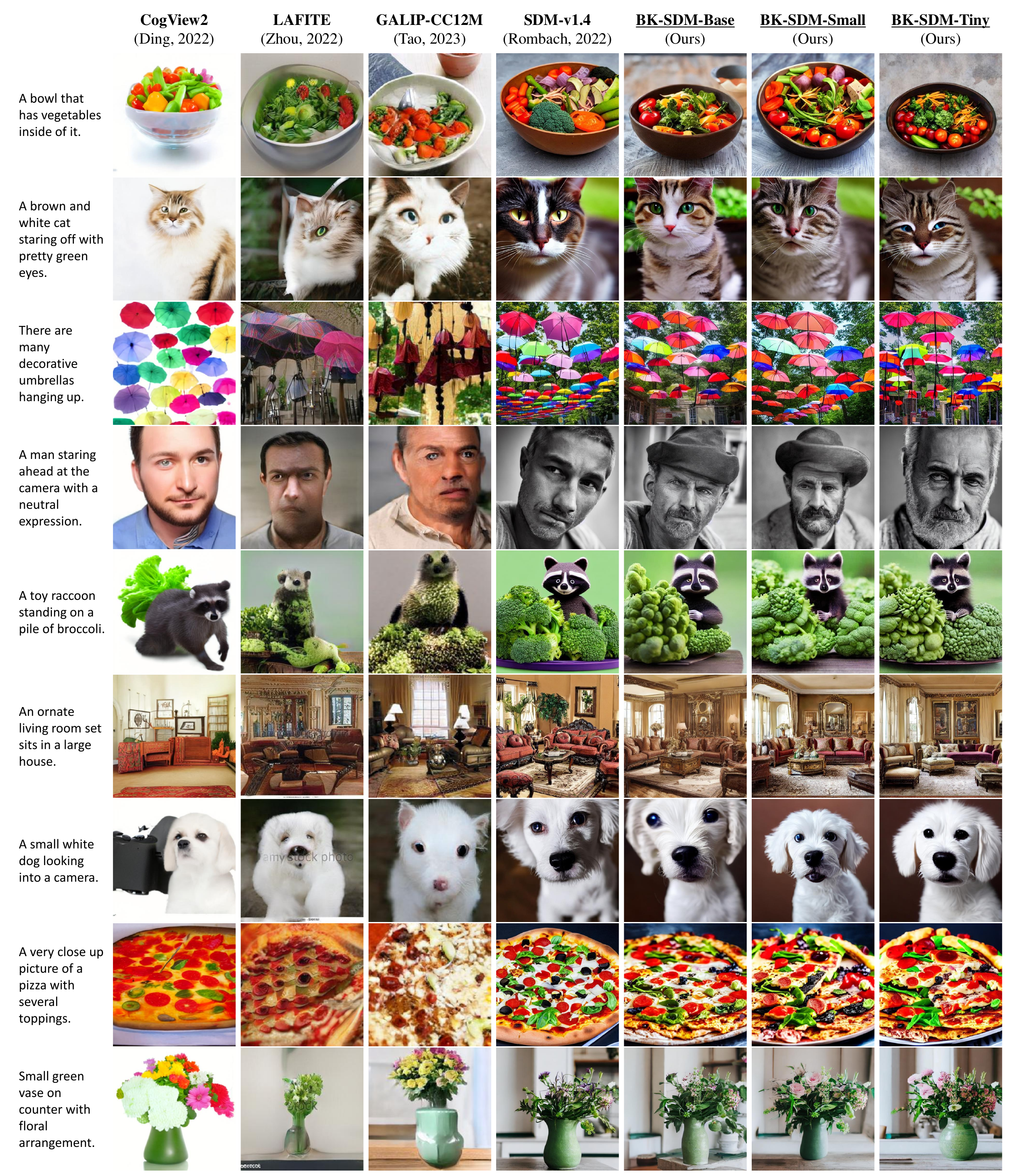}
  \caption{Zero-shot general-purpose T2I results. The results of previous studies~\cite{cogview2,lafite2022,galip} were obtained with their official codes and released models. We do not apply any CLIP-based reranking for SDM and our models.}
  \label{fig_maincomp_supple}
\end{figure*}

\clearpage

\section{Personalized Generation} \label{appendix_dreambooth}
\begin{figure*}[h!]
  \centering
    \includegraphics[width=\linewidth]{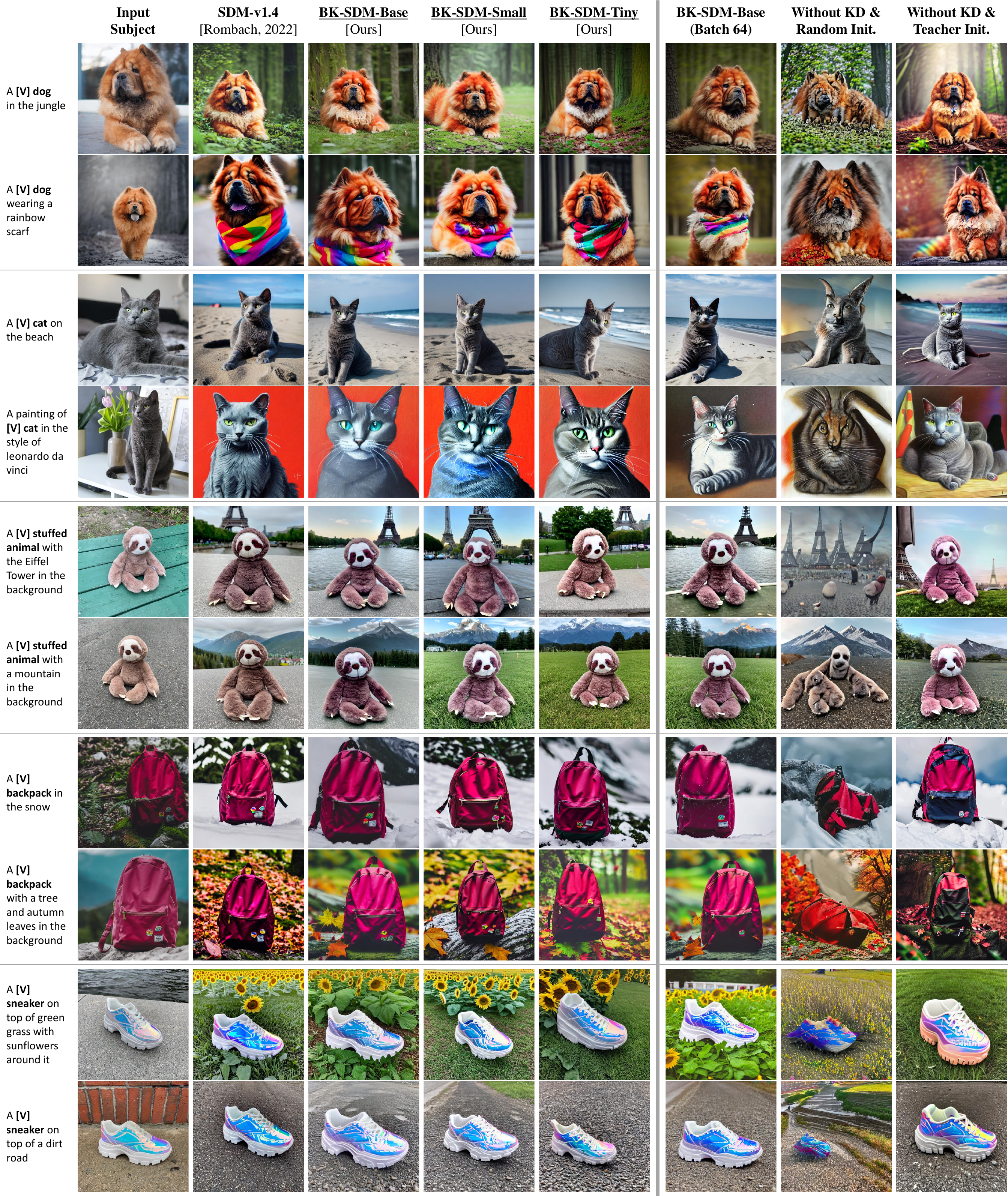}
  \caption{Results of personalized generation. Each subject is marked as ``a [identifier] [class noun]” (e.g., ``a [V] dog"). Similar to the original SDM, our compact models can synthesize the images of input subjects in different backgrounds while preserving their appearance.}
  \label{fig_dreambooth_supple}
\end{figure*}

\clearpage

\section{Text-guided Image-to-Image Translation} \label{appendix:img2img}
\begin{figure*}[h!]
  \centering
    \includegraphics[width=\linewidth]{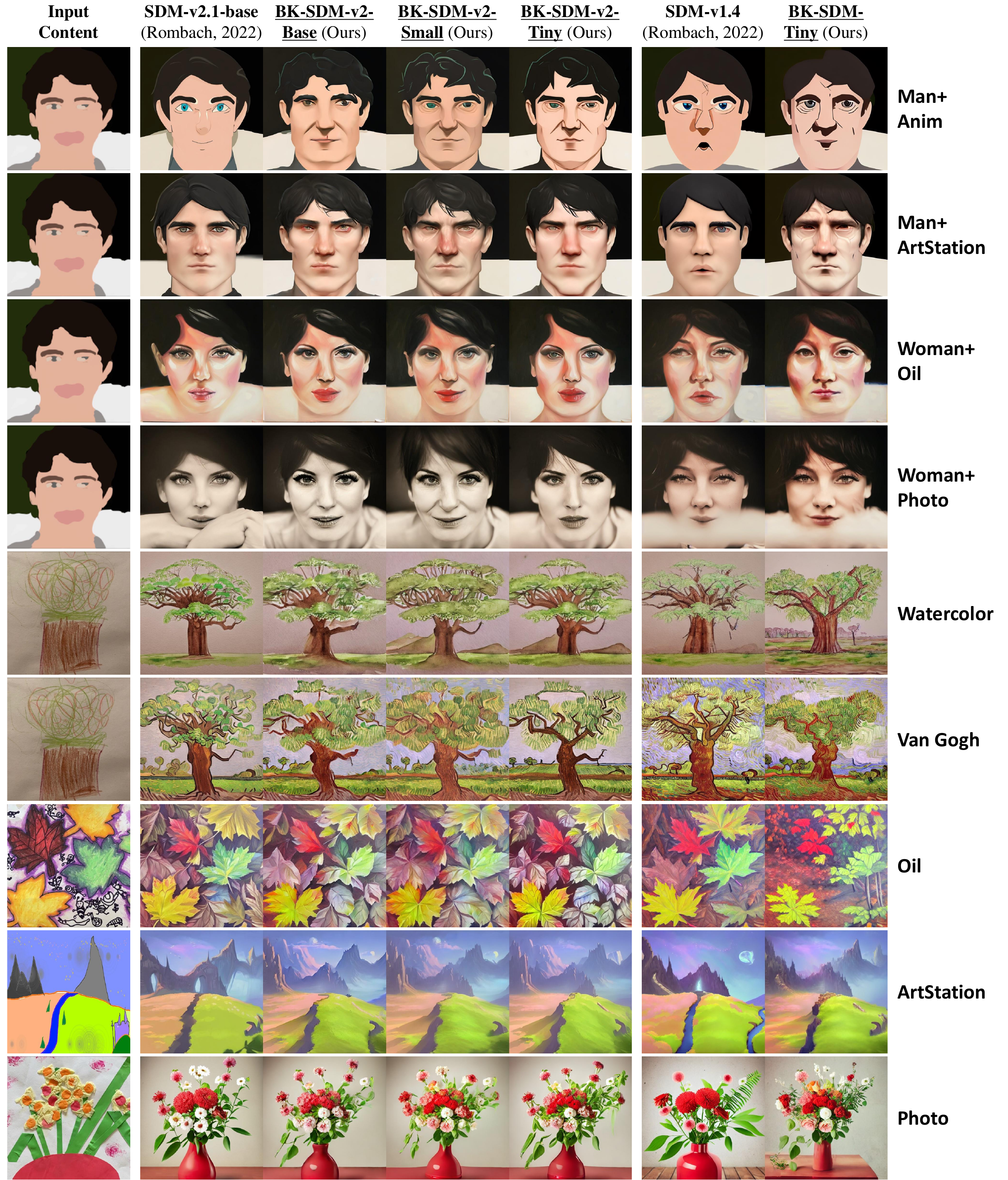}
  \caption{Results of text-guided image-to-image translation. Our small models effectively stylize input images.}
  \label{fig_img2img_supple}
\end{figure*}

\clearpage

\section{Deployment on Edge Devices} \label{appendix_edge_deploy}
\begin{figure*}[h!]
  \centering
    \includegraphics[width=0.95\linewidth]{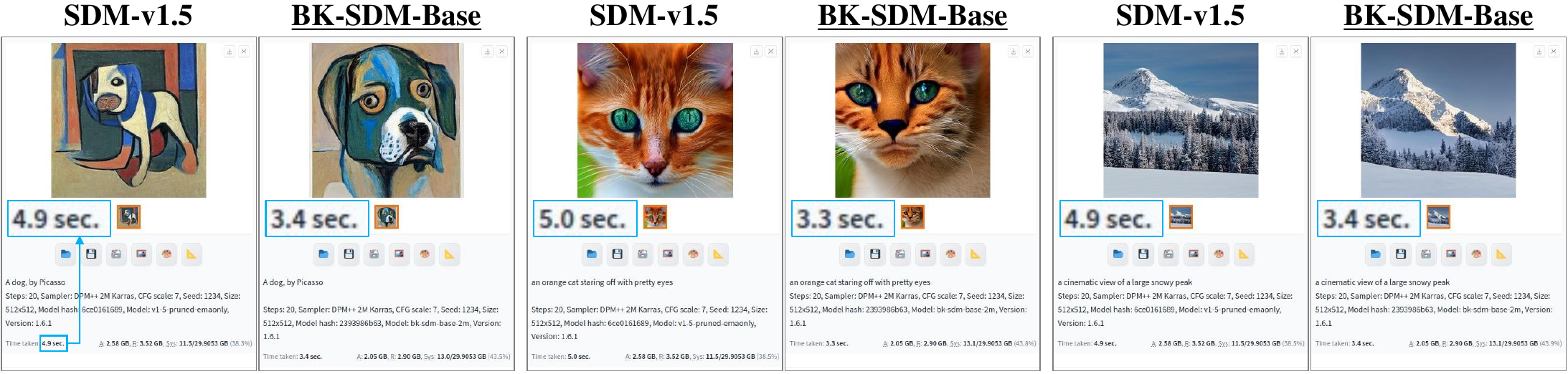}
  \caption{Deployment on NVIDIA Jetson AGX Orin 32GB.}
  \label{fig_supple_orin_demo}
\end{figure*}

\begin{figure*}[h]
  \centering
    \includegraphics[width=\linewidth]{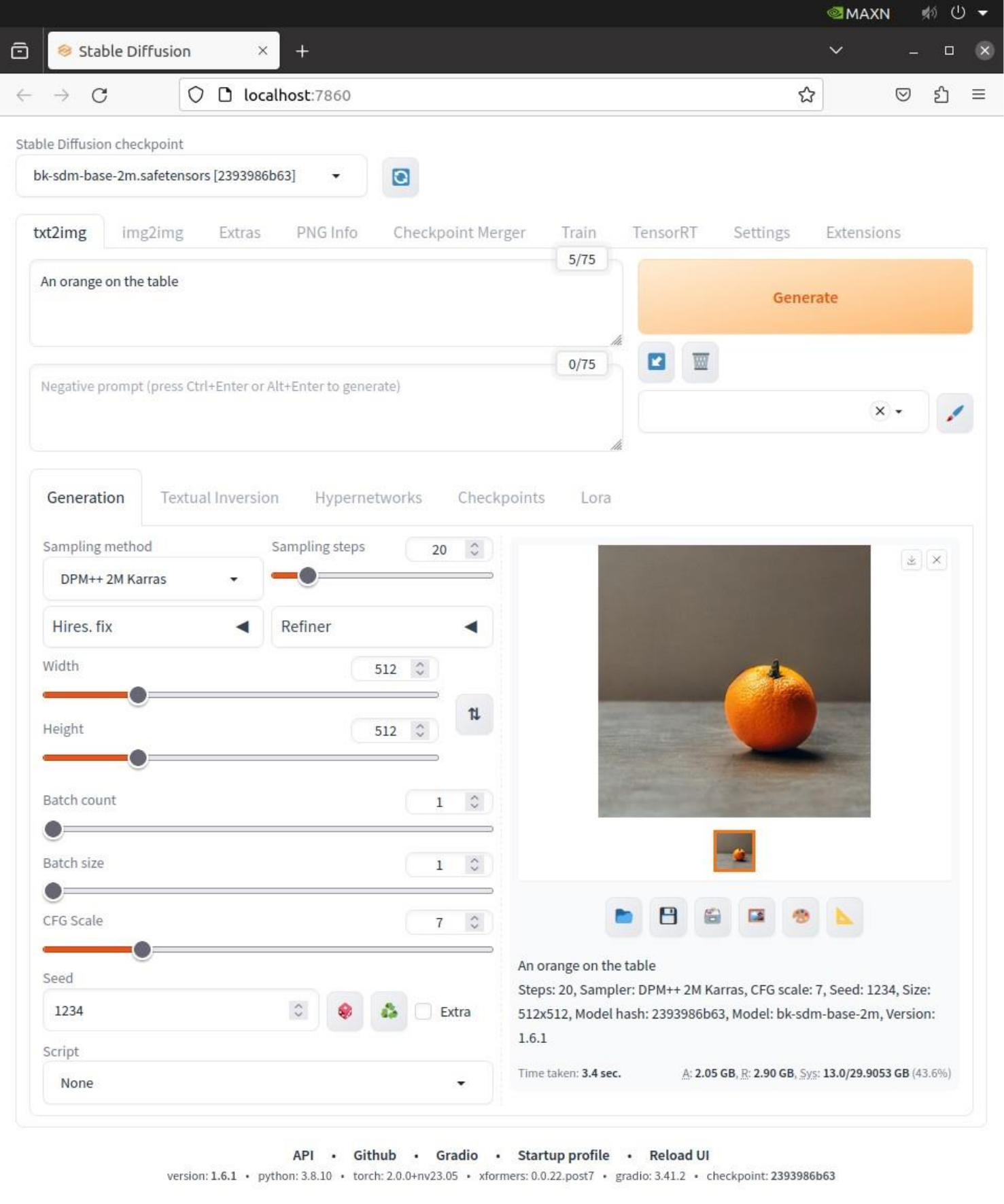}
  \caption{Stable Diffusion WebUI~\cite{sd-webui} used in the deployment on AGX Orin.}
  \label{fig_supple_orin_ui}
\end{figure*}

Our models are tested on NVIDIA Jetson AGX Orin 32GB, benchmarked against SDM-v1.5~\cite{sdm_v1.5_hf,ldm2022} under the same default setting of Stable Diffusion WebUI~\cite{sd-webui}. For the inference, 20 denoising steps, DPM++ 2M Karras sampling~\cite{dpm_solver_2,karras2022elucidating}, and xFormers-optimized attention~\cite{xFormers2022} are used to synthesize 512×512 images. BK-SDM shows quicker generation at 3.4 seconds, compared to the 4.9 seconds of SDM-v1.5 (see Figs.~\ref{fig_supple_orin_demo} and \ref{fig_supple_orin_ui} with BK-SDM-Base trained on 2.3M pairs).

\begin{figure*}[h!]
  \centering
  \includegraphics[width=0.9\linewidth]{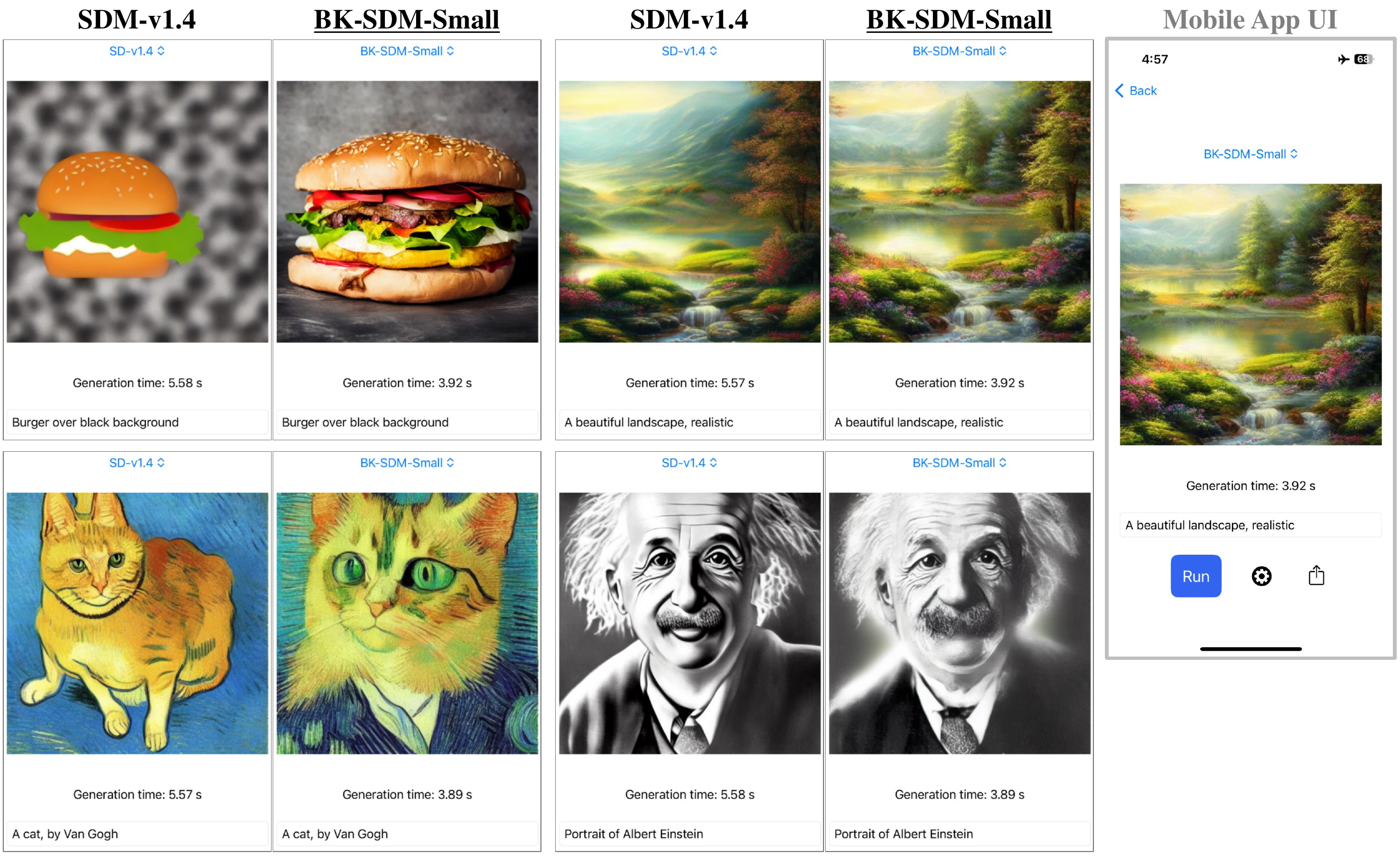}
  \caption{Deployment on iPhone 14.}
  \label{fig_supple_mobile_demo}
\end{figure*}

We also deploy our models on iPhone 14 with post-training palettization~\cite{sd-coreml-apple} and compare them against the original SDM-v1.4~\cite{sdm_v1.4_hf,ldm2022} converted with the identical setup. With 10 denoising steps and DPM-Solver~\cite{dpm_solver_1,dpm_solver_2}, 512×512 images are generated from given prompts. The inference takes 3.9 seconds using BK-SDM, which is faster than 5.6 seconds using SDM-v1.4, while maintaining acceptable image quality (see Fig.~\ref{fig_supple_mobile_demo} with BK-SDM-Small trained on 2.3M pairs).

\begin{figure*}[h]
  \centering
    \includegraphics[width=0.9\linewidth]{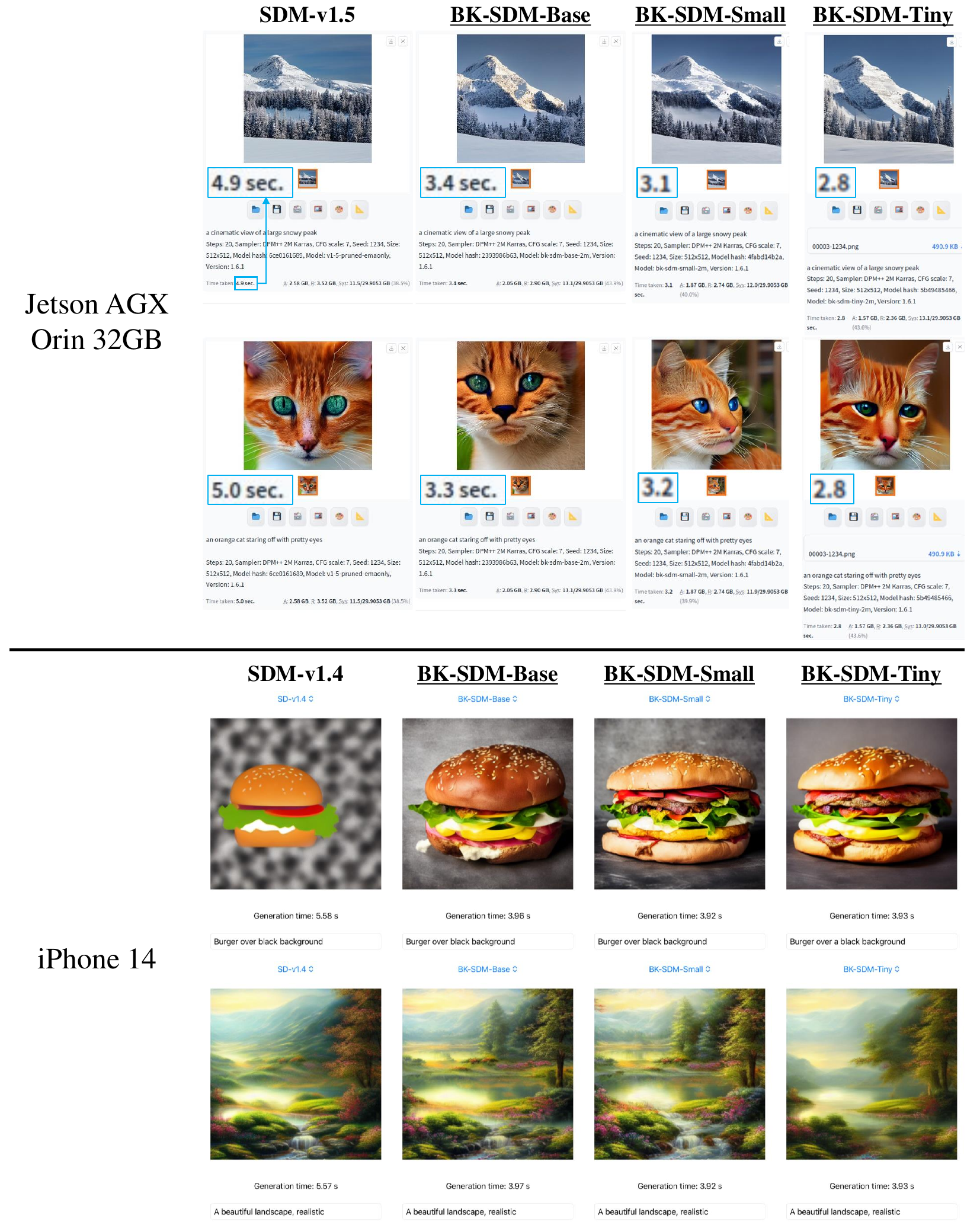}
  \caption{Additional examples from deployment on edge devices.}
  \label{fig_supple_edge_results}
\end{figure*}

\clearpage
Additional results using different models can be found in Fig.~\ref{fig_supple_edge_results}. 

\clearpage
\section{Impact of Training Data Volume}\label{appendix:data_volume}

\begin{figure*}[h]
  \centering
    \includegraphics[width=\linewidth]{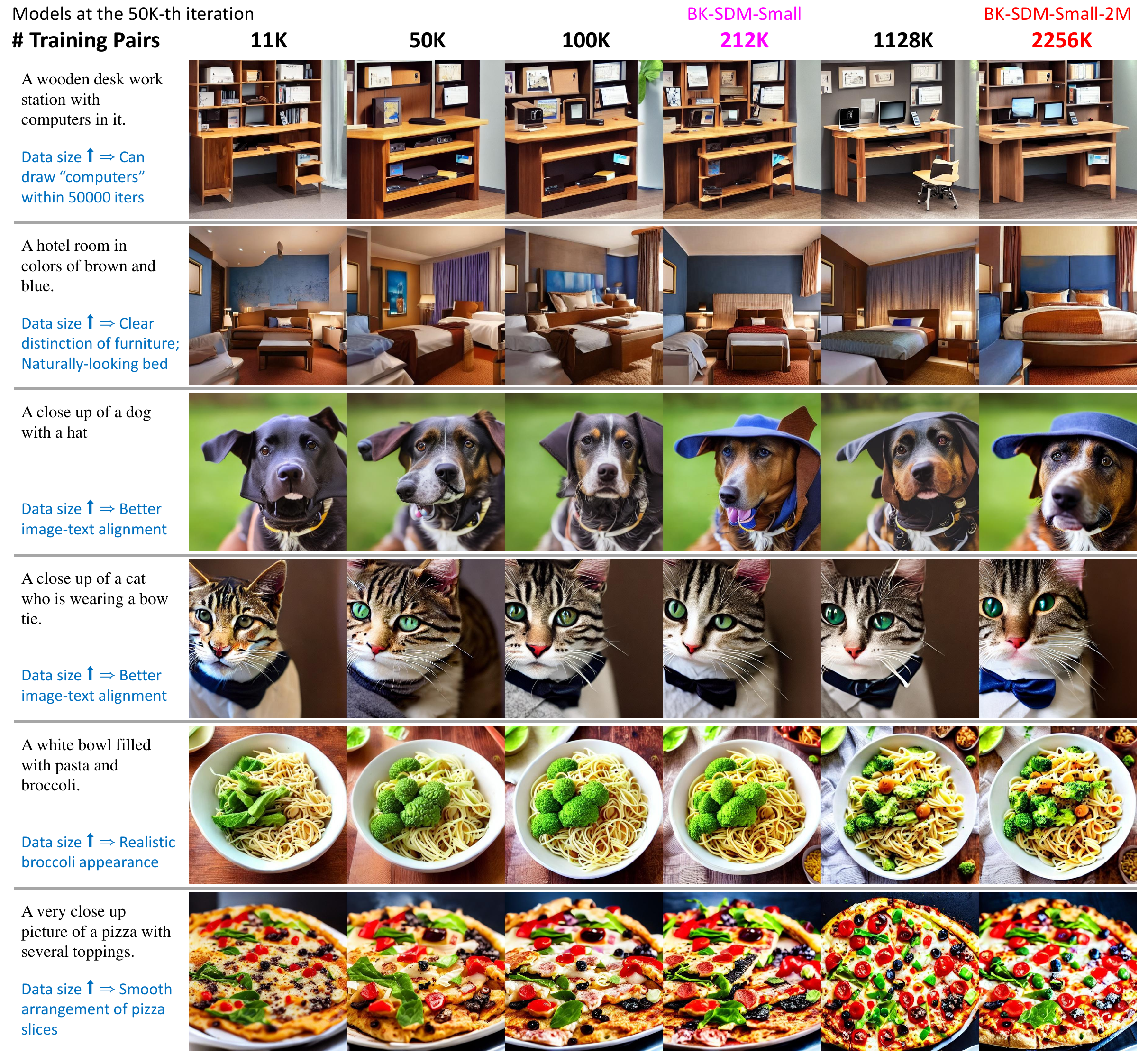}
  \caption{Varying data quantities in training BK-SDM-Small. As the amount of data increases, the visual outcomes improve, such as enhanced image-text matching and clearer differentiation between objects.}
  \label{fig_supple_datasize_bk_small}
\end{figure*}

\begin{figure*}[h]
  \centering
    \includegraphics[width=\linewidth]{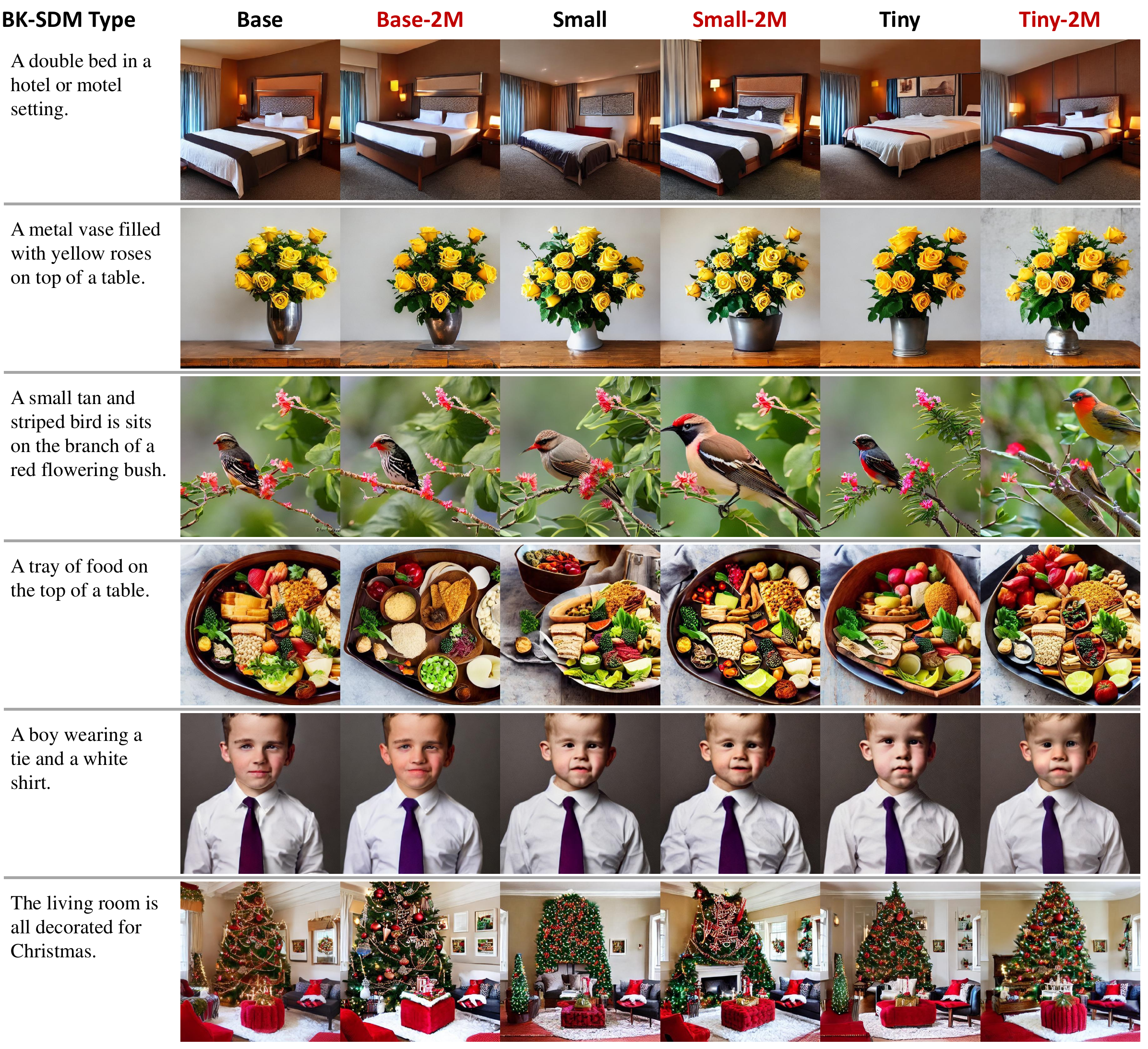}
  \caption{Results of BK-SDM-\{Base, Small, Tiny\} trained on 0.22M pairs and \{Base-2M, Small-2M, Tiny-2M\} trained on 2.3M pairs.}
  \label{fig_supple_2m_models}
\end{figure*}

Fig.~\ref{fig_supple_datasize_bk_small} illustrates how varying data sizes affects the training of BK-SDM-Small. Fig.~\ref{fig_supple_2m_models} presents additional visual outputs of the following models: BK-SDM-\{Base, Small, Tiny\} trained on 212K (i.e., 0.22M) pairs and BK-SDM-\{Base-2M, Small-2M, Tiny-2M\} trained on 2256K (2.3M) pairs.

\clearpage

\clearpage

\section{Additional Experiments}\label{appendix:additional_exp}

\noindent\textbf{More Architectural Exploration.} A model that falls between the original and our base size can be achieved by removing the mid-stage (see Tab.~\ref{appendix_table:midrm_v2.1-base}). Meanwhile, the CLIP criterion can be used to obtain models of various sizes (Tab.~\ref{appendix_table:btw-base-small_sd-v1.4}), though their performance is suboptimal. 

\begin{table}[h]
\centering
\caption{A model between the original and our base size.} \label{appendix_table:midrm_v2.1-base}
\begin{adjustbox}{max width=0.83\columnwidth}
\begin{threeparttable}
\begin{tabular}{c|ccc|c}
\specialrule{.2em}{.1em}{.1em} 

Model                 & \hspace{0.2cm}FID↓\hspace{0.2cm}  & \hspace{0.2cm}IS↑\hspace{0.2cm}   & \hspace{0.2cm}CLIP↑\hspace{0.2cm}  & \hspace{0.2cm} \# Param, U-Net\hspace{0.2cm} \\ \hline
\hspace{0.2cm} Original SD-v2.1-base \hspace{0.2cm} & 13.93 & 35.93 & 0.3075 & 866M     \\
\rowcolor[HTML]{ECF4FF} 
Mid-Stage Removal     & 14.84 & 37.29 & 0.3093 & 769M     \\
Ours-v2-Base        & 15.85 & 31.70 & 0.2868 & 584M     \\ 
\specialrule{.2em}{.1em}{.1em} 
\end{tabular}
\begin{tablenotes}[para,flushleft]
Retraining with batch 128, 0.22M data, 50K iters.
\end{tablenotes}
\end{threeparttable}
\end{adjustbox}
\end{table}

\begin{table}[h]
\centering
\caption{Additional structural variation. The CLIP-Score criterion can be used to yield models of multiple sizes, but their results are inferior to ours.} \label{appendix_table:btw-base-small_sd-v1.4}
\begin{adjustbox}{max width=0.85\columnwidth}
\begin{threeparttable}
\begin{tabular}{cc|ccc|c}
\specialrule{.2em}{.1em}{.1em} 

\multicolumn{2}{c|}{Model (\# Blocks Removed)}                                                 & \hspace{0.2cm}FID↓\hspace{0.2cm}                          & \hspace{0.2cm}IS↑\hspace{0.2cm}                           & \hspace{0.2cm}CLIP↑ \hspace{0.2cm}                         & \hspace{0.2cm}U-Net    \hspace{0.2cm}                    \\ \hline
\multicolumn{1}{c|}{}                             & \hspace{0.2cm} CLIP Criterion (15) \hspace{0.2cm}                                 & 14.06                         & 30.91                         & 0.2787                         & 606M                         \\
\multicolumn{1}{c|}{\multirow{-2}{*}{Base Size}}  & \cellcolor[HTML]{ECF4FF}Ours-v1-Base (14)  & \cellcolor[HTML]{ECF4FF}15.02 & \cellcolor[HTML]{ECF4FF}32.40 & \cellcolor[HTML]{ECF4FF}0.2841 & \cellcolor[HTML]{ECF4FF}580M \\ \hline
\multicolumn{1}{c|}{\hspace{0.2cm} In-Between \hspace{0.2cm}}                   & \hspace{0.2cm} CLIP Criterion (17)  \hspace{0.2cm}                                & 17.65                            & 27.06                            & 0.2553                            & 540M                         \\ \hline
\multicolumn{1}{c|}{}                             & \hspace{0.2cm} CLIP Criterion (19) \hspace{0.2cm}                                 & 21.86                         & 22.01                         & 0.2283                         & 497M                         \\
\multicolumn{1}{c|}{\multirow{-2}{*}{Small Size}} & \cellcolor[HTML]{ECF4FF}Ours-v1-Small (17) & \cellcolor[HTML]{ECF4FF}16.83 & \cellcolor[HTML]{ECF4FF}30.40 & \cellcolor[HTML]{ECF4FF}0.2668 & \cellcolor[HTML]{ECF4FF}483M \\ 
\specialrule{.2em}{.1em}{.1em} 
\end{tabular}
\begin{tablenotes}[para,flushleft]
Retraining with batch 128, 0.22M data, 50K iters.
\end{tablenotes}
\end{threeparttable}
\end{adjustbox}
\end{table}

\noindent\textbf{Effect of Learning Rate (LR).} LRs of 5e-5 (used in the main paper) and 2.5e-5 yield good results (see Tab.~\ref{appendix_table:learn_rate}). Extremely high or low LR values are detrimental.

\begin{table}[h]
\centering
\caption{Effect of learning rate (LR). BK-SDM-v2-Small.} \label{appendix_table:learn_rate}
\begin{adjustbox}{max width=0.65\columnwidth}
\begin{threeparttable}
\begin{tabular}{c|cc
>{\columncolor[HTML]{ECF4FF}}c cc}
\specialrule{.2em}{.1em}{.1em} 

\hspace{0.4cm}LR\hspace{0.4cm}    & \hspace{0.2cm}1.0e-5 \hspace{0.2cm}& \hspace{0.2cm}2.5e-5\hspace{0.2cm} & \hspace{0.2cm}5.0e-5 \hspace{0.2cm}& \hspace{0.2cm} 1.0e-4 \hspace{0.2cm} & \hspace{0.2cm} 2.5e-4 \hspace{0.2cm} \\ \hline
FID↓  & 15.24  & 15.69  & 16.61  & 17.43  & 18.93  \\
IS↑   & 29.77  & 31.64  & 31.73  & 30.58  & 28.90  \\
CLIP↑ & 0.2844 & 0.2906 & 0.2901 & 0.2871 & 0.2775 \\ 

\specialrule{.2em}{.1em}{.1em} 
\end{tabular}
\begin{tablenotes}[para,flushleft]
Retraining with batch 128, 0.22M data, 50K iters.

\end{tablenotes}
\end{threeparttable}
\end{adjustbox}
\end{table}

\begin{figure*}[h]
  \centering
    \includegraphics[width=0.8\linewidth]{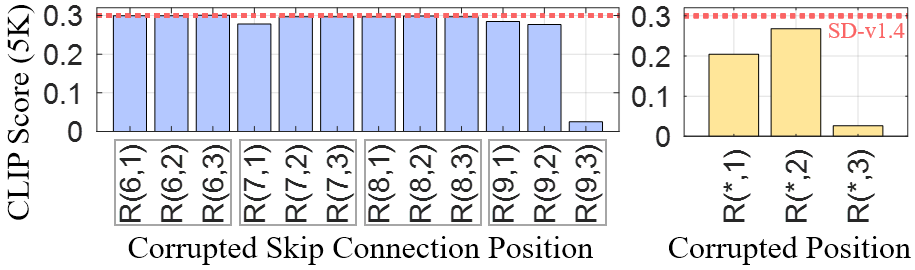}
  \caption{Analysis of skip connections. We corrupt incoming features from channel concatenation in each up stage (left) and multiple stages (right). Higher scores imply removable units.}
  \label{fig:skipanal_zero}
\end{figure*}

\noindent\textbf{Analysis of Skip Connections.} We remove the second channel concatenation (concat) and R-A pairs in each up stage, while retaining the first and third ones. For further analysis, we corrupt the features from skip connections by forcibly assigning zero values (see Fig.~\ref{fig:skipanal_zero}). Consistent with our design, the inner concats are very robust to zeroing and are prunable. Moreover, the second concats are more removable than the others. Note that the first R blocks are often unprunable to utilize the teacher’s weights.

\clearpage

\section{Implementation}\label{appendix:impl_details}

We adjust the codes in Diffusers~\cite{diffusers} for distillation retraining and PEFT~\cite{peft} for per-subject finetuning, both of which adopt the training process of DDPM~\cite{ddpm} in latent spaces. 

\noindent \textbf{Distillation Retraining for General-purpose T2I.} For augmentation, smaller edge of each image is resized to 512, and a center crop of size 512 is applied with random flip. We use a single NVIDIA A100 80G GPU for 50K-iteration retraining with the AdamW optimizer and a constant learning rate of 5e-5. The number of steps for gradient accumulation is always set to 4. With a total batch size of 256 (=4×64), it takes about 300 hours and 53GB GPU memory. Training smaller architectures results in 5$\sim$10\% decrease in GPU memory usage.

\noindent \textbf{DreamBooth Finetuning.} For augmentation, smaller edge of each image is resized to 512, and a random crop of size 512 is applied. We use a single NVIDIA GeForce RTX 3090 GPU to finetune each personalized model for 800 iterations with the AdamW optimizer and a constant learning rate of 1e-6. We jointly finetune the text encoder as well as the U-Net. For each subject, 200 class images are generated by the original SDM. The weight of prior preservation loss is set to 1. With a batch size of 1, the original SDM requires 23GB GPU memory for finetuning, whereas BK-SDMs require 13$\sim$19GB memory.

\noindent \textbf{Inference Setup.} Following the default setup, we use PNDM scheduler~\cite{pndm} for zero-shot T2I generation and DPM-Solver~\cite{dpm_solver_1,dpm_solver_2} for DreamBooth results. For compute efficiency, we always opt for 25 denoising steps of the U-Net, unless specified. The classifier-free guidance scale \cite{clsfreeguide,imagen} is set to the default value of 7.5, except the analysis in Fig.~\ref{fig:tradeoff}.

\noindent \textbf{Image-to-Image Translation.} We use the SDEdit method~\cite{sdedit} implemented in Diffusers~\cite{diffusers}, with the strength value of 0.8.

\noindent \textbf{Distillation Retraining for Unconditional Face Generation.} A similar approach to our T2I training is applied. For the 30K-iteration retraining, we use a batch size of 64 (=4×16) and set the KD loss weights to 100.

\end{document}